\documentclass[12pt]{article}
\usepackage{amsmath, amsthm,amsfonts,amssymb}

\usepackage[round]{natbib}

\usepackage[colorlinks,citecolor=blue,urlcolor=blue]{hyperref}
\usepackage{graphicx,psfrag,epsf}

\usepackage[utf8]{inputenc} %
\usepackage{hyperref}       %
\usepackage{url}            %
\usepackage{booktabs}       %
\usepackage{amsfonts}       %
\usepackage{nicefrac}       %
\usepackage{microtype}      %

\usepackage{mathtools}
\usepackage{graphicx}
\usepackage{enumerate,xspace}
\usepackage{amsfonts}
\usepackage{amssymb}
\usepackage{fancyhdr}
\usepackage{comment}
\usepackage{parskip}
\usepackage{float}
\usepackage{longtable}
\usepackage{tablefootnote}
\usepackage{url}
\usepackage{standalone}
\restylefloat{table}
\usepackage{mathrsfs}

\newtheorem{remark}{Remark}

\usepackage{tikz}
\usetikzlibrary{matrix,arrows,calc,shapes,backgrounds}
\usetikzlibrary{shapes.callouts,decorations.text} 
\usetikzlibrary{shapes.misc}

\newcommand{\calD}{{\mathcal{D}}}
\newcommand{\calE}{{\mathcal{E}}}

\newcommand{\calG}{{\mathcal{G}}}
\newcommand{\calH}{{\mathcal{H}}}
\newcommand{\calI}{{\mathcal{I}}}

\newcommand{\calL}{{\mathcal{L}}}

\newcommand{\calN}{{\mathcal{N}}}
\newcommand{\calO}{{\mathcal{O}}}
\newcommand{\calP}{{\mathcal{P}}}

\newcommand{\calR}{{\mathcal{R}}}
\newcommand{\calS}{{\mathcal{S}}}

\newcommand{\calX}{{\mathcal{X}}}

\DeclarePairedDelimiterX{\infdivx}[2]{(}{)}{%
	#1\;\delimsize\|\;#2%
}
\newcommand{\infdiv}{D\infdivx}

\DeclarePairedDelimiterX{\infdivxL}[2]{\big(}{\big)}{%
	#1\;\delimsize\|\;#2%
}

\usepackage{prettyref,xspace}
\newrefformat{eq}{\eqref{#1}}
\newrefformat{thm}{Theorem~\ref{#1}}
\newrefformat{fact}{Fact~\ref{#1}}
\newrefformat{sec}{Section~\ref{#1}}
\newrefformat{algo}{Algorithm~\ref{#1}}
\newrefformat{fig}{Fig.~\ref{#1}}
\newrefformat{tab}{Table~\ref{#1}}
\newrefformat{rmk}{Remark~\ref{#1}}
\newrefformat{def}{Definition~\ref{#1}}
\newrefformat{cor}{Corollary~\ref{#1}}
\newrefformat{lmm}{Lemma~\ref{#1}}
\newrefformat{prop}{Proposition~\ref{#1}}
\newrefformat{supp}{Supplement~\ref{#1}}
\newrefformat{app}{Appendix~\ref{#1}}
\newrefformat{ex}{Example~\ref{#1}}
\newrefformat{ass}{Assumption~\ref{#1}}

\DeclareMathOperator*{\argmax}{arg\,max}

\def\bt{\mathrm{t}}
\def\bx{\mathbf{x}}

\def\bz{\mathbf{z}}

\def\bbeta{\boldsymbol{\beta}}

\def\bTheta{\boldsymbol{\Theta}}

\newtheorem{axiom}{Axiom}

\newtheorem{theorem}{Theorem}[section]
\newtheorem{lemma}[theorem]{Lemma}
\newtheorem{corollary}[theorem]{Corollary}
\newtheorem{assumption}[axiom]{Assumption}
\newtheorem{definition}[theorem]{Definition}

\usepackage{stmaryrd}

\usepackage{xcolor}

\mathtoolsset{showonlyrefs}
\usepackage{xr-hyper}

\makeatletter
\long\def\XR@test#1#2#3#4\XR@{%
  \ifx#1\newlabel
     \expandafter\protected@xdef\csname r@\XR@prefix#2\endcsname
       {\XR@addURL{#3}}%
     \expandafter\protected@xdef\csname isxr@r@\XR@prefix#2\endcsname{1}%
  \else\ifx#1\@input
     \edef\XR@list{\XR@list\filename@area#2\relax}%
  \fi\fi
  \ifeof\@inputcheck\expandafter\XR@aux
  \else\expandafter\XR@read\fi}

\AtBeginDocument{%
    \let\orig@T@ref\T@ref
    \def\T@ref#1{%
        \ifcsname isxr@r@#1\endcsname
            \@refstar{#1}%
        \else
            \orig@T@ref{#1}%
        \fi
    }%
}%
\makeatother
\newcommand{\blind}{1}

\usepackage[margin=1in]{geometry}

\def\spacingset#1{\renewcommand{\baselinestretch}%
{#1}\small\normalsize} \spacingset{1}

\begin{document}

\date{}

\if1\blind
{
  \title{\bf Large Scale Prediction with Decision Trees}
  \author{Jason M. Klusowski\thanks{Supported in part by the National Science Foundation through grants DMS-2054808 and HDR TRIPODS DATA-INSPIRE DCCF-1934924.}\hspace{.2cm}
  \and Peter M. Tian\thanks{Supported in
part by the Gordon Wu Fellowship of Princeton University.} \\
    Department of Operations Research and Financial Engineering, \\
Princeton University}
   \maketitle
} \fi

\if0\blind
{
  \bigskip
  \bigskip
  \bigskip
  \begin{center}
    {\LARGE\bf Large Scale Prediction with Decision Trees}
\end{center}
  \medskip
} \fi

\bigskip
\begin{abstract}
This paper shows that decision trees constructed with Classification and Regression Trees (CART) and C4.5 methodology are consistent for regression and classification tasks, even when the number of predictor variables grows sub-exponentially with the sample size, under natural 0-norm and 1-norm sparsity constraints. The theory applies to a wide range of models, including (ordinary or logistic) additive regression models with component functions that are continuous, of bounded variation, or, more generally, Borel measurable. Consistency holds for arbitrary joint distributions of the predictor variables, thereby accommodating continuous, discrete, and/or dependent data. Finally, we show that these qualitative properties of individual trees are inherited by Breiman’s random forests. A key step in the analysis is the establishment of an oracle inequality, which allows for a precise characterization of the goodness-of-fit and complexity tradeoff for a mis-specified model.
\end{abstract}

\noindent%
{\it Keywords:}  CART; C4.5; random forests; interpretable machine learning; oracle inequality
\vfill

\newpage
%\spacingset{1.59} % DON'T change the spacing!

\section{Introduction}

Decision trees are one of the most elemental methods for predictive modeling. Accordingly, they are the cornerstone of many celebrated algorithms in statistical learning. For example, decision trees are often employed in ensemble learning, i.e., bagging \citep{breiman1996bagging}, random forests \citep{breiman2001random}, and gradient tree boosting \citep{friedman2001greedy}. 
From an applied perspective, decision trees scale well to large datasets, and are intuitive and interpretable, the latter of which makes them easy to explain to statistical non-experts, particularly in the context of rule-based decision-making. They are also supplemented by a rich set of analytic and visual diagnostic tools for exploratory data analysis. These qualities have led to the prominence of decision trees in disciplines---such as medicine and business---which place high importance on the ability to understand and interpret the output from
the training algorithm, even at the expense of predictive accuracy.\footnote{The oft-touted interpretability of CART and C4.5 is sometimes compromised by the instability of the splits, particularly among deeper nodes where less data is available. Therefore one should be cautious when attaching any meaning or interpretation to such splits.}

Though our primary focus is theoretical, to make this paper likewise relevant to the applied user of decision trees, we focus on Classification And Regression Trees (CART) \citep{breiman1984} and C4.5 \citep{quinlan1993c4} methodology---undoubtedly the most popular varieties for regression and classification problems. On the theoretical side, these approaches raise a number of technical challenges which stem from the top down greedy recursive splitting and line search needed to find the best split points, thereby making CART and C4.5 notoriously difficult to study. These subtle mechanisms are of course desirable from a statistical standpoint, as they endow the decision tree with the ability to adapt to structural and qualitative properties of the underlying statistical
model (such as sparsity and smoothness). 
Notwithstanding these major challenges, we take a significant step forward in advancing the theory of decision trees and prove the following (informal) statement in this paper:
\begin{quotation}
\noindent\normalsize\textit{Decision trees constructed with CART and C4.5 methodology are consistent for large scale predictive models, where the number of predictor variables is allowed to grow sub-exponentially with the sample size, under $ \ell_0 $ or $\ell_1$ sparsity constraints.}
\end{quotation}
The consistency (with respect to mean squared error risk for regression and excess mis-classification risk for classification) is shown under essentially no assumptions on the predictor variables, thereby improving upon most past work which requires the them to be continuous and either independent or near-independent (e.g., uniformly distributed or with joint densities which are bounded above and below by fixed positive constants). 

Expectedly, our results for individual trees also carry over to ensembles, namely, Breiman's random forests \citep{breiman2001random}, which among other things, use CART methodology for the constituent trees.

\subsection{Prior art}
We now review some of the past theoretical work on decision trees, starting with CART in the regression setting, and then for C4.5 in the classification setting.

\paragraph{Regression trees.} The first consistency result for CART was provided in the original book that proposed the methodology \citep{breiman1984}, albeit under very strong assumptions on the tree construction, such as a minimum node size condition and shrinking cell condition. 
Thirty years later, \cite{scornet2015consistency} showed asymptotic consistency of CART for (fixed dimensional) additive regression models with continuous component functions, en route to establishing asymptotic consistency of Breiman's random forests. This paper was an important technical breakthrough because it did not require any of the strong assumptions on the tree made in \cite{breiman1984}. Subsequent work by \cite{chi2020asymptotic, klusowski2020sparse, syrgkanis2020estimation, wager2015} provide finite sample consistency rates in a high dimensional setting with exact sparsity, though again, like \cite{breiman1984}, they operate under a set of conditions that may or may not hold in practice. Another notable paper by \cite{gey2005model} provides oracle-type inequalities for pruned CART, but the theory does not extend to out-of-sample prediction.

Motivated by Stone's conditions for consistency in non-parametric regression \citep{stone1977consistent}, most existing convergence results for decision trees follow an approach in which the approximation error is bounded by the mesh of the induced partition of the input space. Conditions are then imposed, either explicitly or implicitly, to ensure that the mesh approaches zero as the depth of the tree increases. This is then combined with a standard empirical process argument to show vanishing estimation error, which in turn, implies that the prediction risk vanishes also \citep{denil2014, breiman1984, wager2015, wager2018estimation}. 
In contrast, the aforementioned paper \citep{scornet2015consistency} controls the variation of the regression function inside the cells of the partition, without explicitly controlling the mesh, though the theoretical consequences are similar. While these techniques can be useful to prove consistency statements, they are not generally delicate enough to capture the adaptive properties of the tree or handle high dimensional situations. 

More recently, \cite{chi2020asymptotic, klusowski2020sparse, syrgkanis2020estimation} developed techniques to directly analyze the approximation error (instead of using the granularity of the partition as a proxy) by exploiting the greedy optimization inherent in CART methodology. These papers provide consistency rates for models with exact sparsity in a high dimensional regime (i.e., when the ambient dimensionality grows with the sample size); however, they make a number of assumptions that lead to an unsatisfactory theory. For example, the results of \cite{klusowski2020sparse} apply only to the noise free setting, \cite{chi2020asymptotic} require the ambient dimensionality to grow at most polynomially with the sample size,
and \cite{syrgkanis2020estimation} work with binary valued predictor variables. In addition, a local accuracy gain condition (akin to an edge or progress condition in boosting literature) is required in these works \citep{chi2020asymptotic, klusowski2020sparse, syrgkanis2020estimation} to ensure the approximation error decreases by a constant factor after splitting at each level. In \cite{chi2020asymptotic}, this local accuracy gain condition is verified to hold for some simple classes of additive regression models, such as those with isotonic or piece-wise linear component functions and independent predictor variables. \cite{syrgkanis2020estimation} do not provide any concrete examples of models that satisfy what they call a \emph{sub-modularity} property, and the reader is required to accept its validity. In summary, it is difficult to verify which models satisfy these technical conditions and, therefore, the general applicability of the theory remains unclear. 

\paragraph{Classification trees.} The story for the classification setting is far less complete. Theory for regression trees is far easier to assemble—and has therefore been overwhelmingly the focus of past literature—since one does not have to deal with the discrete nature of the model. This lack of attention is somewhat unfortunate as decision trees are more often successfully deployed in problems with discrete outputs, i.e., clinical decision support systems. As an exception, one stand-alone paper to tackle the classification problem is \cite{kearns1999boosting}, where the authors show that classification trees constructed with CART 
 and C4.5 methodology have small mis-classification risk under a weak hypothesis assumption, i.e., the decision tree output in each node performs slightly better than random guessing as approximations to the target function. Their results, however, do not account for the effect of dimensionality, nor do they accommodate standard statistical models for classification, such as logistic regression.

 \subsection{Related work} \label{sec:related}
 
In closing, we mention a few papers that study other tree based procedures, but with different aims and from different perspectives. Mondrian random forests \citep{mourtada2020minimax, mourtada2021amf}, unlike some of the aforementioned variants of Breiman's random forests, provably attain near-optimal minimax rates for various levels of smoothness regularity. The dyadic CART procedure of \citet{donoho1997cart}, obtained by optimal (non-greedy), dyadic recursive partitioning, was also shown to achieve near-optimal rates (and adaptation to unknown smoothness regularity) for the case of two predictor variables. (Dyadic CART unfortunately scales poorly with the dimensionality $p$, since finding the optimal dyadic partition may require up to $ \calO(2^p N p) $ operations \citep{chatterjee2021}.)
An interesting line of work from a Bayesian perspective explores Bayesian trees and forests. For example, \citet{rockova2020posterior, jeong2020art} obtain near-optimal posterior concentration rates for non-parametric regression, similar to the optimal minimax rates discussed in \prettyref{sec:consistent_discussion} below, that adapt to both exact sparsity and smoothness regularity (for Bayesian CART and BART) and additive structures (for BART).

\subsection{Organization}

This paper is organized according the following schema. In \prettyref{sec:learning}, we describe a unified statistical framework for regression and classification problems, and introduce various important quantities for performance assessment.
We review basic terminology associated with decision trees and describe CART and C4.5 methodology in \prettyref{sec:cart}. Our main results for CART and C4.5 are contained in \prettyref{sec:main}; specifically, an empirical risk bound, oracle inequality, and asymptotic consistency statement for (ordinary or logistic) additive regression models.
We then show that C4.5 achieves a faster consistency rate for separable, large margin binary data in \prettyref{sec:margin}. Our main consistency results in \prettyref{sec:main} are extended to models with interactions in \prettyref{sec:beyond_additive} and Breiman's random forests in \prettyref{sec:forests}. We conclude with a discussion in \prettyref{sec:discussion}.
Finally, all proofs and technical lemmas are contained in the supplementary material.

	\section{Statistical framework} \label{sec:learning}
	
	Throughout this paper, we operate under a standard predictive framework for regression and classification; i.e., from training data, we desire to predict a response value $ y $ for a new set of $ p $ predictor variables $ \bx $. More formally, we observe training data $ \calD \coloneqq \{(\bx_1, y_1), (\bx_2, y_2), \dots, (\bx_N, y_N)\} $ drawn i.i.d. from the statistical model $ \mathbb{P}_{(\bx, y)} = \mathbb{P}_{\bx}\mathbb{P}_{y\mid\bx} $, where $ \mathbb{P}_{\bx} $ is a probability measure on the $ \sigma$-algebra of Borel subsets of $ \mathbb{R}^p $.
	To understand the predictive properties of decision trees in an (ultra) high dimensional setting, 
	the dimensionality $ p = p_N$ is permitted to grow sub-exponentially with the sample size $N$.

	In order to unify
both the regression and classification settings, we consider a discriminative statistical model in which the conditional mean of $y$ given $\bx$ is modeled indirectly via a (possibly non-linear) link function.
	To this end, we assume 
	that there is a fixed and known link function $h: \mathbb{R} \rightarrow \mathbb{R}$ such that 
	\begin{equation} \label{eq:g^*}
		g^*(\bx) \coloneqq h(\mathbb{E}(y \mid \bx))  \in \calG,
	\end{equation}
	where $\calG$ is some class of $\mathbb{R}^p \rightarrow \mathbb{R}$ functions to be chosen ahead.
 A loss function $\calL(\cdot,\cdot)$ is chosen so that
	\begin{equation} \label{eq:pop_min}
		g^*(\cdot) \in {\rm argmin}_{g(\cdot) \in \calG }\mathbb{E}_{(\bx, y)} \calL(y, g(\bx)).
	\end{equation}
	The \emph{empirical and population risk} corresponding to the loss function $\calL(\cdot,\cdot)$ are denoted by
	\begin{equation} \label{eq:pop_loss}
	\widehat \calR(g) \coloneqq \frac{1}N\sum_{i=1}^N \calL(y_i,g(\bx_i)), \qquad \calR(g) \coloneqq \mathbb{E}_{(\bx, y)} \calL(y, g(\bx)),
	\end{equation}
	respectively.
	In light of \prettyref{eq:pop_min}, we aim to estimate the true model $g^*(\cdot)$ by finding a $\calD$-dependent fit $g(\cdot)$, not necessarily in $ \calG $, which
	 comes close to minimizing the empirical risk $\widehat \calR(g)$.

	\subsection{Regression setting}\label{sec:regression}
	We assume a real-valued response variable $ y \in \mathbb{R} $ and consider the identity link function 
		$ h(\mu) = \mu $
	so that 
	$
	g^*(\bx) =\mathbb{E}(y \mid \bx)
	$
	is the conditional mean response.
	We choose the loss function to be the squared error 
	\begin{equation}\label{eq:training_loss_r}
		\calL(y, g(\bx)) = (y-g(\bx))^2,
	\end{equation}
	which satisfies \prettyref{eq:pop_min}.
	The out-of-sample performance is measured
	with the squared $ \mathscr{L}_2(\mathbb{P}_{\bx}) $ norm
\begin{equation} \label{eq:L2}
\|g-g^*\|^2 \coloneqq \mathbb{E}_{\bx}((g(\bx) - g^*(\bx))^2),
\end{equation}
which also equals the excess squared error risk $ \calR(g) - \calR(g^*) $ under these model specifications.

	\subsection{Classification setting}\label{sec:classification}
	We assume a binary response variable $y \in \{-1,1\}$ 
	and consider the logit link function 
	$
		h(\mu) = \log(\frac{1+\mu}{1-\mu}),
        $
	so that
	$
		g^*(\bx) = h(2\eta^*(\bx) -1)= {\rm log} (\frac{\eta^*(\bx)}{1-\eta^*(\bx)})
        $
	is the log-odds of the conditional class probability $\eta^*(\bx) \coloneqq \mathbb{P}(y=1\mid\bx) = 1/(1+\exp(-g^*(\bx)))$. Note that if $ g^*(\cdot) $ is a linear function, then this model is linear logistic regression.
	We choose the loss function to be the logistic loss
	\begin{equation} \label{eq:training_loss_c}
		\calL(y, g(\bx)) = \log(1+\exp(-y g(\bx))),
	\end{equation}
	which satisfies \prettyref{eq:pop_min} and, appealingly, corresponds to maximum likelihood estimation of conditionally Bernoulli distributed data with probability model $ \mathbb{P}(y \mid \bx)  = 1/(1+\exp(- yg(\bx))) $.
	Instead of assessing performance with the logistic risk, we more familiarly consider the mis-classification risk 
	\begin{equation} \label{eq:mis-classification} 
	{\rm Err}(g) \coloneqq \mathbb{P}_{(\bx,y)}(c(\bx) \neq y)
	\end{equation}
	 of the plug-in classifier 
	$$
	 c(\bx) \coloneqq 
	\begin{cases}
	+1 & \text{if} \; g(\bx) \geq 0 \\
	-1 & \text{if} \; g(\bx) < 0
	\end{cases}.
	$$
	The out-of-sample performance 
	is then measured by the excess mis-classification risk
	\begin{equation} \label{eq:excess_Err}
	{\rm Err}(g) - {\rm Err}(g^*),
	\end{equation}
	where $ {\rm Err}(g^*) $ is the Bayes error rate.	%
	
\begin{definition} \label{def:prediction_risk}
When no context is provided, \emph{prediction risk} will refer to excess squared error risk \prettyref{eq:L2} in the regression setting (\prettyref{sec:regression}) or excess mis-classification risk \prettyref{eq:excess_Err} in the classification setting (\prettyref{sec:classification}).
\end{definition}

	\section{CART and C4.5 methodology} \label{sec:cart}
	
	As mentioned earlier, regression trees are commonly constructed with Classification and Regression Trees (CART) methodology. In the context of classification, one can use either CART or its contemporary counterpart, C4.5.
	For technical reasons that will be explained at the end of this section,
	 we will focus on the latter methodology. 
	While CART and C4.5 are algorithmically similar, they differ in important ways. The notable distinction lies in the criterion used to determine the split points, which turns out to be the key to their success for the predictive models we consider in \prettyref{sec:regression} and \prettyref{sec:classification}.
	In a nutshell, the objective of decision tree learning is to find partitions of the predictor variables that produce minimal empirical risk of the constant (average response) values over the partition.  
	Because of the computational infeasibility of choosing the best overall partition, CART and C4.5 operate in a greedy, top down fashion (with a mere $ \calO(pN\log^2(N)) $ average-case %
	 complexity \citep[Section 5]{louppe2014understanding}) using a procedure in which a sequence of locally optimal splits recursively partition the input space. 
	 
	 In \prettyref{sec:impurity}, we first discuss the splitting rule, stopping criterion, and tree output for a generic (top down, greedy) tree construction algorithm.
	We then specialize our treatment of decision trees to CART and C4.5 in \prettyref{sec:CART} and \prettyref{sec:C4.5}, respectively.

	\subsection{Greedy tree construction} \label{sec:impurity}
	Consider splitting a decision tree $T$ at a node $\bt$ (a hyperrectangular region in $\mathbb{R}^p$). Let $s$ be a candidate split point for a variable $x_j \in\mathbb{R}$ that divides the parent node $\bt$ into left and right daughter nodes $\bt_L$ and $\bt_R$ according to whether $ x_j \leq s $ or $ x_j > s $, respectively. These two nodes will be denoted by $\bt_L \coloneqq \{\bx \in \bt: x_j \leq s\} $ and $\bt_R \coloneqq \{\bx \in \bt: x_j > s\} $.

	An effective split divides the data from the parent node into two daughter nodes so that the heterogeneity in each of the daughter nodes, as measured through the \emph{impurity}, 
	is maximally reduced from that of the parent node. The impurity is determined by the within-node empirical risk
	\begin{equation} \label{eq:training_loss_node}
		\widehat \calR_\bt(g) \coloneqq \frac{1}{N_\bt}\sum_{\bx_i \in \bt} \calL(y_i, g(\bx_i)), \quad N_\bt \coloneqq \#\{\bx_i \in \bt\}.
	\end{equation}
	In accordance with the directive of minimizing the empirical risk, it is equal to the smallest within-node empirical risk over all constant predictors in the node, i.e., 
	\begin{equation} \label{eq:impurity_loss}
		\calI(\bt)= \min_{\beta \in \mathbb{R}} \widehat \calR_{\bt}(\beta) = \widehat \calR_{\bt}(h(\overline{y}_\bt)), \quad \overline y_{\bt}  \coloneqq \frac{1}{N_\bt}\sum_{\bx_i \in \bt}y_i.
	\end{equation}
	The parent node $\bt$ is split into two daughter nodes using a variable $x_{j_\bt}$ and split point $s_{\bt}$ which produce the largest impurity gain
	\citep[Definition 8.13]{breiman1984},
	 \citep[Page 22]{quinlan1993c4}
	\begin{equation} \label{eq:IG} 
		\mathcal{IG}(j,s, \bt) \coloneqq
		 \calI(\bt)- P_{\bt_L}  \calI(\bt_L)-P_{\bt_R}   \calI(\bt_R), \quad
		 \mathcal{IG}(\bt) \coloneqq 
				\max_{(j,s)} \mathcal{IG}(j,s,\bt),
	\end{equation}
	breaking ties arbitrarily, where $P_{\bt_L} \coloneqq N_{\bt_L}/N_{\bt} $ and $P_{\bt_R} \coloneqq N_{\bt_R}/N_{\bt} $ are the proportions of data points
within $\bt$ that are contained in $\bt_L$ and $\bt_R$, respectively. 
Equivalently, the variable and split point, chosen to maximize \prettyref{eq:IG}, also minimize the within-node empirical risk \prettyref{eq:training_loss_node} over all \emph{decision stumps}, since
	\begin{equation}\label{eq:training_decrease}
	P_{\bt_L}  \calI(\bt_L)+P_{\bt_R}   \calI(\bt_R)= \min_{\beta_0, \beta_1 \in \mathbb{R}} \widehat \calR_\bt(\beta_0+\beta_1\mathbf{1}(x_j > s)).
	\end{equation}
We can thus view the maximum impurity gain $\mathcal{IG}(\bt) $ as the amount by which the optimal decision stump decreases the empirical risk in the node.

The daughter nodes $ \bt_L $ and $ \bt_R $ of $ \bt $ become new parent nodes at the next level of the tree and are themselves further divided according to the previous scheme, and so on and so forth, until a desired depth $K$ is reached.
There are many criteria that can be used to determine when to stop splitting, each one giving rise to a different tree structure. In this paper, we use the following stopping rule.
	\begin{definition}[Stopping rule] \label{def:stopping}
	We stop splitting a node if (i) the node contains a single data point, (ii) all input values and/or all response values within the node are the same, or (iii) a depth of $ K $ is reached, whichever occurs sooner.
	\end{definition}
Finally, the output of the tree $T$ at a terminal node $ \bt $ is the best constant predictor in the node:
\begin{equation} \label{eq:output}
\hat g(T)(\bx) \coloneqq h(\overline{y}_\bt) \approx g^*(\bx), \quad \bx \in \bt.
\end{equation}
When we wish to emphasize the dependence of the tree $ T $ on the depth $ K $, we will write $ T_K $.

	\subsection{CART algorithm} \label{sec:CART}
	Recall the regression setting in \prettyref{sec:regression},
	where $h(\cdot)$ is the identity link and $\calL(\cdot,\cdot)$ is the squared error loss \prettyref{eq:training_loss_r}. In this case, the impurity \prettyref{eq:impurity_loss} becomes the within-node sample variance of the response variable, i.e., 
	\begin{equation} \label{eq:impurity_r}
	\calI(\bt)= \frac{1}{N_\bt }\sum_{\bx_i \in \bt}(y_i - \overline y_{\bt})^2.
	\end{equation} 
	 According to these model specifications, the tree output \prettyref{eq:output} makes a prediction by returning the within-node sample mean of the response variable, i.e.,
	\begin{equation}\label{eq:h_r}
	\hat g(T^{\textnormal{\scalebox{0.8}{CART}}})(\bx) =h(\overline y_{\bt}) = \overline y_{\bt}, \quad \bx \in \bt.
	\end{equation}
	
	\subsection{C4.5 algorithm} \label{sec:C4.5}
	Recall the classification setting in \prettyref{sec:classification}, 
	where $h(\cdot)$ is the logit link and $\calL(\cdot,\cdot)$ is the logistic loss \prettyref{eq:training_loss_c}. In this case, 
	the impurity \prettyref{eq:impurity_loss} becomes the binary entropy of the within-node empirical class probability, i.e.,
		\begin{equation} \label{eq:impurity_c}
		\calI(\bt) = 
		\eta_\bt\log(1/\eta_\bt) + (1-\eta_\bt)\log(1/(1-\eta_\bt)), \quad  \eta_\bt \coloneqq \frac{1}{N_\bt}\sum_{\bx_i \in \bt} \mathbf{1}(y_i = 1).
	\end{equation}
	As entropy quantifies the information content of a random variable, the impurity gain \prettyref{eq:IG} is sometimes called the \emph{information gain} or the \emph{mutual information}.
	According to these model specifications, the tree output \prettyref{eq:output} is the log-odds of the within-node empirical class probability
	\begin{equation} \label{eq:h_c}
	\hat g(T^{\textnormal{\scalebox{0.8}{C4.5}}})(\bx) = h(\overline{y}_\bt) = \log\Big( \frac{\eta_\bt}{1-\eta_\bt} \Big), \quad \bx \in \bt.
	\end{equation}
The tree makes a class prediction by returning the majority vote of the classes in the node, i.e.,
	\begin{equation*} 
	 \hat c(T^{\textnormal{\scalebox{0.8}{C4.5}}})(\bx) = 
	\begin{cases}
	+1 & \text{if} \; \hat g(T^{\textnormal{\scalebox{0.8}{C4.5}}})(\bx) \geq 0 \\
	-1 & \text{if} \; \hat g(T^{\textnormal{\scalebox{0.8}{C4.5}}})(\bx) < 0
	\end{cases}.
	\end{equation*}

		\begin{definition} \label{def:tree}
	A decision tree $T$ constructed with CART methodology (\prettyref{sec:CART}) in the regression setting (\prettyref{sec:regression}) is denoted by $T^{\textnormal{\scalebox{0.8}{CART}}}$.
	Similarly, 
	a decision tree $T$ constructed with C4.5 methodology (\prettyref{sec:C4.5}) in the classification setting (\prettyref{sec:classification}) is denoted by $T^{\textnormal{\scalebox{0.8}{C4.5}}}$.
	An arbitrary unnamed decision tree $T$
	refers to either $T^{\textnormal{\scalebox{0.8}{CART}}}$ or $T^{\textnormal{\scalebox{0.8}{C4.5}}}$. 
\end{definition}	

	\begin{remark} \label{rmk:gini}
		The curious reader may wonder why we do not analyze 
		 classification trees constructed with CART methodology—which use the so-called \emph{Gini} splitting criterion—and instead focus on C4.5 methodology. Gini impurity for classification trees is equivalent to squared error impurity \prettyref{eq:impurity_r} for regression trees \citep[Section 3]{louppe2014understanding}, and thus both types of trees produce identical estimates of the conditional mean response, namely, $\overline y_{\bt} \approx \mathbb{E}(y \mid \bx)  $ for $ \bx \in \bt $. However, our forthcoming results for regression trees are relegated to additive $ \mathbb{E}(y \mid \bx) $, which are appropriate for regression, but awkward for classification.
		For this reason, C4.5 methodology allows us to work with more common models for discrete responses $y\in \{-1, 1\}$, such as the logistic regression model in \prettyref{sec:classification}.
	\end{remark}

	\section{Main results} \label{sec:main}

	In this section, we first describe 
	a class of large scale predictive models.
	For this class of models, we establish an adaptive prediction risk bound, which in turn, leads to consistency of CART and C4.5.

	\subsection{Large scale predictive models} \label{sec:large}

	We illustrate the high dimensional properties of CART and C4.5 in the context of \emph{generalized additive models}, whereby an \emph{additive function} is related to the conditional mean of the response variable by a link function \prettyref{eq:g^*}. While there are many link functions that could be used, the canonical choices for regression and classification tasks are the aforementioned identity and logit functions from \prettyref{sec:regression} and \prettyref{sec:classification}, respectively.
	
	In particular, note that additive logistic regression models are, importantly, different from ordinary additive regression in the sense that
		$ \mathbb{E}(y  \mid \bx) = 2\mathbb{P}(y = 1 \mid \bx)-1 $ is not equal to an additive function of the predictor variables. This means that one cannot deduce consistency
from existing results for regression trees \citep{scornet2015consistency}, even in the fixed dimensional setting, since they are limited to additive $ \mathbb{E}(y \mid \bx) $. 

In practice, additive logistic regression models are typically fit with backfitting or boosting algorithms \citep{hastie1990generalized, tutz2006generalized}. Despite a rich literature on theoretical guarantees in the fixed dimensional setting \citep{horowitz2004nonparametric}, consistency results in the (ultra) high dimensional setting (i.e., $ \log(p) \asymp N^{1-\xi} $, $ \xi \in (0, 1) $) with either $ \ell_0 $ or $ \ell_1 $ sparsity constraints do not appear to be available, unless
	 the logistic regression model is linear and the sparsity pattern is the number of relevant predictor variables \citep{abramovich2019high}. 
	Therefore, sparse logistic regression models render a situation in which decision trees are unrivaled as a scalable, theoretically grounded method.

\paragraph{Generalized additive models.}	We now describe the generalized additive modeling framework with additional precision. Consider the additive function class
	$$
	\calG^1 \coloneqq \big\{ g(\bx) \coloneqq g_1(x_1) + g_2(x_2) + \cdots + g_p(x_p) \big\},
	$$
	where $ g_1(x_1), g_2(x_2), \dots, g_p(x_p) $ is a collection of $ p $ univariate (Borel measurable) functions.
	The generalized additive modeling framework involves finding a $g(\cdot) \in \calG^1$ for which
	\begin{equation} \label{eq:additive}
		 g(\bx) = g_1(x_1) + g_2(x_2) + \cdots + g_p(x_p)
	\end{equation}
	approximates the true model \prettyref{eq:g^*}.

Generalized additive models are often used in high dimensional settings, in part because notions of exact $ \ell_0 $ or approximate $ \ell_1 $ sparsity are easy to define and interpret. We now describe these two types of sparsity patterns in detail.

\paragraph{Approximate $ \ell_1 $ sparsity.}	As we have already mentioned, we would like to consider models with \emph{approximate} sparsity. To this end, for $g(\cdot) \in \calG^1$, we define the \emph{total variation $\ell_1 $ norm} $ \|g\|_{{\rm TV}} $ as the infimum of 
	\begin{equation} \label{eq:TV_sum}
	{\rm TV}(g_1) + {\rm TV}(g_2) + \cdots + {\rm TV}(g_p)
	\end{equation}
	over all representations of $ g(\cdot) $ as \prettyref{eq:additive}, i.e., $ \|g\|_{{\rm TV}} $ is the aggregated total variation of the individual component functions (see \cite{tan2019doubly} and the references therein).
	To simplify the arguments, we henceforth assume $ g(\cdot) $ has a canonical representation \prettyref{eq:additive} such that \prettyref{eq:TV_sum} achieves this infimum. 
One can think of $ \|g\|_{{\rm TV}} $ as a measure of the capacity of $ g(\cdot) $ and, as we shall see, it will play a central role in the paper. The total variation $ \ell_1 $ norm is a desirable quantification of sparsity because it allows for some predictor variables to make very small yet meaningful contributions to the model.

	In the case that all the component functions $ g_j(\cdot) $ are smooth over a domain $ \calX $ with Lebesgue measure one, the total variation $ \ell_1 $ norm can be expressed as the multiple Riemann integral
	$$
	\|g\|_{{\rm TV}} = \int_{\calX^p} \|\nabla g(\bx)\|_{\ell_1}d\bx,
	$$
	where $ \nabla(\cdot) $ is the gradient operator and $ \|\cdot\|_{\ell_1} $ is the usual $ \ell_1 $ norm of a vector in $ \mathbb{R}^p $.
	In particular, if $ g(\bx) = \bbeta^T \bx $ is linear over $\calX^p$, then $ \|g\|_{{\rm TV}} = \|\bbeta\|_{\ell_1} $, the $ \ell_1 $ norm of the coefficient vector $ \bbeta \in \mathbb{R}^p $.

\paragraph{Exact $ \ell_0 $ sparsity.} 	To account for \emph{exact} sparsity, we also define the $ \ell_0 $ norm $  \|g\|_{\ell_0} $ of $g(\cdot) \in \calG^1$ as the infimum of
	$$ \#\{ j : g_j(\cdot) \; \text{is non-constant} \}$$
	over all representations of $ g(\cdot) $ as \prettyref{eq:additive}.
	In other words, the $ \ell_0 $ norm counts the number of relevant variables that affect $ g(\cdot) $. For $g(\cdot) \in\calG^1$,
	we have the relation
	$$
	\|g\|_{{\rm TV}} \leq \|g\|_{\ell_0} \cdot \max_j {\rm TV}(g_j),
	$$
	provided $ \max_j {\rm TV}(g_j) < \infty $.
	Thus, a small total variation  $\ell_1$ norm captures \emph{either} exact or approximate sparsity, whichever is present. 
	
	Finally, our results will require us to have uniform control on the magnitude of a function $ g : \mathbb{R}^p \to \mathbb{R} $, which we do through the supremum norm $ \|g\|_{\infty} \coloneqq \sup_{\bx} |g(\bx)| $.

	\subsection{Empirical risk bound} \label{sec:training}
	
	The empirical risk bound in this section is the key to all forthcoming results.
	We prove a purely algorithmic guarantee, namely, that the (excess) empirical risk of a depth $ K $ regression tree constructed with CART or C4.5 methodology is of order $ 1/K$. To the best of our knowledge, this result is the first of its kind for any decision tree algorithm. The math behind it is surprisingly simple; in particular, unlike most past work, we do not need to directly analyze the partition of the input space that is induced by recursively splitting the variables. Nor do we need to rely on concentration of measure to show that certain local (i.e., within-node) empirical quantities concentrate around their population level versions. Because we are able to circumvent these technical aspects with a new method of analysis, the astute reader will notice and appreciate that we make no assumptions on the decision tree itself (such as a minimum node size condition or shrinking cell condition that typifies extant literature). In contrast with the recent work of \cite{chi2020asymptotic, syrgkanis2020estimation}, we also do not need to assume a local accuracy gain condition so that the approximation error decreases by a constant factor after splitting at each level. 
	
	We now describe the empirical risk bound in detail. We aim to upper bound the excess empirical risk 
	of the decision tree $T_K$. In view of \prettyref{eq:training_decrease},
	as we grow deeper trees, the empirical risk reduces by the impurity gain, as can be seen from the recursion (which holds generically)
	\begin{equation} \label{eq:training_recursion}
		\widehat \calR(\hat g(T_K)) = \widehat \calR(\hat g(T_{K-1})) - \sum_{\bt \in T_{K-1}} \frac{N_\bt}N   \mathcal{IG}(\bt),
	\end{equation}
	where the notation ``$ \bt \in T$'' means that $ \bt $ is a terminal node of of a tree $T$. 
	In view of \prettyref{eq:training_recursion}, to obtain an inductive upper bound on $\widehat \calR(\hat g(T_K))$, we further aim to the lower bound the impurity gain $  \mathcal{IG}(\bt) $ for $\bt \in T_{K-1}$ in terms of the within-node excess risk $ \widehat \calR_{\bt}(\hat g(T_{K-1})) - \widehat \calR_{\bt}(g) $ for a candidate model $ g(\cdot) $. \prettyref{lmm:gain} below accomplishes this goal.

	\begin{lemma}[Impurity gain for CART and C4.5] \label{lmm:gain}
		Let $g(\cdot) \in \calG^1$ be any additive function and $K\ge 1$ be any depth.
		Then for any terminal node $ \bt $ of the tree $T_{K-1}$ such that $ \widehat \calR_\bt(\hat g(T_{K-1})) > \widehat \calR_\bt (g) $,
	 we have
		$$
		\mathcal{IG}(\bt)\geq \frac{(\widehat \calR_\bt(\hat g(T_{K-1})) - \widehat \calR_\bt (g))^2}{V^2(g)},
		$$
		where
		$V(g) = \|g\|_{\rm TV}$ for CART and $V(g) = \|g\|_{\rm TV}+\|g\|_{\infty} + 3$ for C4.5.
	\end{lemma}

	Plugging \prettyref{lmm:gain} into \prettyref{eq:training_recursion} and subtracting $ \widehat \calR(g) $ from both sides, we see that 
	\begin{equation} \label{eq:training_recursion_E}
		\calE_K \leq \calE_{K-1}- \frac{1}{V^2(g)}\sum_{\bt\in T_{K-1}:\calE_{K-1}(\bt) > 0}\frac{N_\bt}N \calE^2_{K-1}(\bt),
	\end{equation}
	where
	\begin{equation} \label{eq:calE}
	\calE_K \coloneqq \widehat \calR(\hat g(T_K)) - \widehat \calR(g), \quad \calE_K(\bt) \coloneqq \widehat \calR_\bt(\hat g(T_K)) - \widehat \calR_\bt(g)
	\end{equation}
	are the global and within-node excess empirical risks, respectively.
	Next, using Jensen's inequality on \prettyref{eq:training_recursion_E} and the fact that $ \calE_{K-1} = \sum_{\bt\in T_{K-1}}\frac{N_\bt}N\calE_{K-1}(\bt) $,
	it can be shown that (see \prettyref{lmm:gaind} in \prettyref{supp:non-additive})
	\begin{equation} \label{eq:main_recursion}
		\calE_K\leq \calE_{K-1}\Big(1- \frac{\calE_{K-1}}{V^2(g)}\Big), \quad \calE_{K-1} \geq 0, \quad K \geq 1.
	\end{equation}
	Iterating the recursion \prettyref{eq:main_recursion},
	 we establish the following upper bound on the excess empirical risk $\calE_K$.

	\begin{theorem}[Empirical risk bound for CART and C4.5]
		\label{thm:training}
		Let $T_K$ be a depth $K\geq 1$ decision tree.
		 Then we have
		$$
		\widehat \calR(\hat g(T_K))\leq \inf_{g(\cdot)\in\calG^1} \Big\{\widehat \calR(g) + 
		\frac{V^2(g)}{K+3}
		\Big\},
		$$
		where $V(g)$ is the constant specified in \prettyref{lmm:gain}.
	\end{theorem}
	
	The above theorem says that a decision tree of depth $ K $ minimizes the empirical risk (for squared error loss and logistic loss) over all additive functions, up to a slackness term of order $ 1/K$.

	\subsection{Oracle inequality} \label{sec:oracle}

	Our main theorem
	establishes an adaptive prediction risk bound (also known as an \emph{oracle inequality}) for decision trees under model mis-specification; that is, when the true model \prettyref{eq:g^*} may not belong to $ \calG^1 $. Essentially, the result says that CART and C4.5 adapt to the class of (ordinary or logistic) additive regression models, performing as if they were finding the best additive approximation of the true model \prettyref{eq:g^*},
	 while accounting for the capacity (the total variation $ \ell_1 $ norm $ \|\cdot\|_{{\rm TV}} $) of the approximation.

	In the regression setting, for simplicity and ease of exposition, we assume that the error $\varepsilon = y- g^*(\bx) = y - \mathbb{E}(y \mid \bx)$ is sub-Gaussian, i.e., there exists $ \sigma^2 > 0 $ such that for all $ u \geq 0 $,
	\begin{equation} \label{eq:bounded}
		\mathbb{P}(|\varepsilon| \geq u) \leq 2\exp(-u^2/(2\sigma^2)).
	\end{equation}
	Before we state our main theorem, we remind the reader that $\|\cdot \|$ denotes the $ \mathscr{L}_2(\mathbb{P}_{\bx}) $ norm
	 \prettyref{eq:L2} and ${\rm Err}(\cdot)$ denotes the mis-classification risk \prettyref{eq:mis-classification}. Using these risk measures, we evaluate the performance of CART for the regression model in \prettyref{sec:regression} and the performance of C4.5 for the logistic regression model in \prettyref{sec:classification}.

	\begin{theorem}[Oracle inequalities for CART and C4.5]\label{thm:oracle}
		Let $K\geq 1$ be any depth. Granting the noise condition 
		 \prettyref{eq:bounded}, we have
		\begin{equation} \label{eq:oracle}
			\mathbb{E}_\calD(\|\hat g(T_K^{\textnormal{\scalebox{0.8}{CART}}}) - g^*\|^2) \leq  
			2\inf_{g(\cdot)\in\calG^1} \Bigg\{ \|g - g^*\|^2 + \frac{\|g\|^2_{{\rm TV}}}{K+3} + C_1\frac{2^K\log^2(N)\log(Np)}N  \Bigg\},
		\end{equation}
		where $ C_1$ is a positive constant that depends only on $ \|g^*\|_\infty $ and $ \sigma^2 $.
		Furthermore, 
		we have
		\begin{equation}\label{eq:oracle_c}
		\begin{aligned}
		&\mathbb{E}_{\calD}({\rm Err}(\hat g(T_{K}^{\textnormal{\scalebox{0.8}{C4.5}}} ))) - {\rm Err}(g^*) \\
			& \qquad \leq 
			\inf_{g(\cdot)\in\calG^1}  \Bigg\{ \|g-g^*\|+ 2\frac{\|g\|_{{\rm TV}}+\|g\|_\infty+3}{\sqrt{K+3}} +C_2\Big(\frac{2^K\log^2(N)\log(Np)}{N}\Big)^{1/4}\Bigg\},
		\end{aligned}
		\end{equation}
		where 
		$C_2$ is a positive universal constant.
	\end{theorem}
\begin{remark}
	Throughout this paper, we measure the accuracy of a predictor via the \emph{expected} prediction risk over the data $ \calD $, as in \prettyref{thm:oracle}. However, at the expense of more complicated expressions, one can also obtain statements that hold with high probability.
\end{remark}
	\prettyref{thm:oracle} reveals the tradeoff between the goodness-of-fit and complexity relative to sample size. The goodness-of-fit terms involving $ \|g - g^*\| $, $ \|g\|_{{\rm TV}}$, and $ \|g\|_{\infty} $ stem from the excess empirical risk bound in \prettyref{thm:training}, and the descriptive complexity term $ 2^K\log(Np) $ comes from the fact that the empirical $\epsilon$-metric entropy for depth $ K $ decision trees with $ p $ predictor variables is of order $ 2^K\log(Np/\epsilon) $.

	\subsection{Consistency} \label{sec:consistent_rate}
	We now explore the case where the true model $g^*(\cdot)$ has the freedom to change with the sample size, which is common in other literature on high dimensional consistency
	\citep{buhlmann2006boosting}.
	To this end, \prettyref{thm:oracle} immediately implies
	 consistency when the model is well-specified (i.e., $g^*(\cdot) \in \calG^1$) and has a controlled sparsity pattern.
	More specifically, choosing $g(\cdot) = g^*(\cdot)$ in \prettyref{thm:oracle} and stipulating that
	 the total variation $ \ell_1 $ norm of $ g^*(\cdot)$
	 does not grow too fast, we find that CART and C4.5 are consistent, even when the dimensionality grows sub-exponentially with the sample size. We note that this type of result is impossible with non-adaptive procedures that do not automatically adjust the amount of smoothing along a particular dimension according to how much the predictor variable affects the response variable. Such procedures perform local estimation at a query point using data that are close in every single dimension, making them prone to the curse of dimensionality even if the true model is sparse. This is the case with conventional multivariate (Nadaraya-Watson or local polynomial) kernel regression in which the bandwidth is the same for all directions, or $ k $-nearest neighbors with Euclidean distance. Indeed, one can compute asymptotic expansions of their bias and variance \citep{ruppert1994multivariate, mack1981local}, which evidently do not exploit low dimensional structure in the regression model. 
	 
	\begin{corollary}[Consistency of CART and C4.5] \label{cor:consistent}
		Consider a sequence of prediction problems \prettyref{eq:g^*} with true models $\{g^*_N(\cdot)\}_{N=1}^\infty$.
		Assume that $g^*_N(\bx) = \sum_{j=1}^{p_N} g_{j}(x_j) \in \calG^1$ and $ \sup_N \|g^*_N\|_{\infty} < \infty $. Suppose that $ K_N \rightarrow \infty $, $ \|g_N^*\|_{{\rm TV}} = o(\sqrt{K_N})$, and $\frac{2^{K_N}\log^2(N)\log(Np_N)}N \rightarrow 0$ as $N \rightarrow \infty$.
		Granting the noise condition \prettyref{eq:bounded}, regression trees are consistent, that is,
		$$
		\lim_{N \rightarrow\infty} \mathbb{E}_\calD(\| \hat g(T_{K_N}^{\textnormal{\scalebox{0.8}{CART}}}) - g^*_N\|^2) = 0.  
		$$
Furthermore, classification trees are consistent, that is,
		$$
		\lim_{N \rightarrow\infty} (\mathbb{E}_{\calD}({\rm Err}(\hat g(T_{K_N}^{\textnormal{\scalebox{0.8}{C4.5}}}))) - {\rm Err}(g^*_N))
		= 0.
		$$
	\end{corollary}

	\begin{remark} \label{rmk:exact}
		Note that because $ \|g^*_N\|_{{\rm TV}} \leq \|g^*_N\|_{\ell_0} \cdot \max_{j\le p_N} {\rm TV}(g^*_{j}) $, the consistency statement in \prettyref{cor:consistent} also applies to models with sparsity patterns defined by the number of relevant variables. Thus, the condition $  \|g_N^*\|_{{\rm TV}} = o(\sqrt{K_N}) $ can be replaced with $ \|g_N^*\|_{\ell_0} = o(\sqrt{K_N}) $, provided $ \max_{j\le p_N} {\rm TV}(g^*_{j}) $ is independent of $ N$. The same reasoning applies to all forthcoming results that use $ \|\cdot\|_{{\rm TV}} $ to measure sparsity.
	\end{remark}
	
\subsubsection{Consistency rates} 
We now describe the effect of specific choices of the depth and regimes of the ambient dimension on the consistency rate for CART and C4.5. The hypotheses of \prettyref{cor:consistent} are satisfied if, for example, $ K_N = \lfloor (\xi/2)\log_2(N) \rfloor$, $ \log(p_N) \asymp N^{1-\xi} $ for $ \xi \in (0, 1) $, $\sup_N\|g_N^*\|_{\infty} < \infty$, and $\sup_N\|g_N^*\|_{\rm TV} < \infty$.
In this case, from \prettyref{eq:oracle} in \prettyref{thm:oracle}, the consistency rate of the CART algorithm is 
\begin{equation} \label{eq:rate}
\frac{4\sup_N\|g_N^*\|^2_{{\rm TV}}}{\xi\log_2(N)+6} 
+ 2C_1\frac{\log^3(N)}{N^{1 - \xi/2}} + 2C_1\frac{\log^2(N)}{N^{\xi/2}}
= \calO(1/\log(N))
\end{equation}
and from \prettyref{eq:oracle_c} in \prettyref{thm:oracle}, 
the consistency rate of the C4.5 algorithm is
\begin{equation} \label{eq:rate_logistic}
4\frac{\sup_N\|g_N^*\|_{{\rm TV}}+\sup_N\|g_N^*\|_{{\infty}}+3}{\sqrt{2\xi\log_2(N)+12}} 
+ C_2\Big(\frac{\log^3(N)}{N^{1 - \xi/2}} +\frac{\log^2(N)}{N^{\xi/2}}\Big)^{1/4} =	 \calO(1/\sqrt{\log(N)}).
\end{equation}
The dependence on the total variation $\ell_1$ norm $ \|g_N^*\|_{{\rm TV}} $ in the consistency rates \prettyref{eq:rate} and \prettyref{eq:rate_logistic} shows that CART and C4.5 can tolerate an approximate sparsity level that grows as fast as $ o(\sqrt{\log(N)}) $. As per \prettyref{rmk:exact}, a similar growth is tolerated for the $ \ell_0 $ norm $ \|g_N^*\|_{\ell_0} $.

\subsubsection{Consistency for unbounded variation component functions} \label{sec:unbounded}

\prettyref{cor:consistent} implicitly considers (ordinary or logistic) additive regression models whose component functions
have bounded variation,
 per the finiteness of $ \|g_N^*\|_{\text{TV}} $.
In fact,
consistency holds when the component functions $ g_j(\cdot) $ are merely Borel measurable,
as 
\prettyref{cor:consistent_general} reveals.

\begin{corollary}[Consistency of CART and C4.5 for unbounded variation components] \label{cor:consistent_general}
	Consider a sequence of prediction problems with true models $\{g^*_N(\cdot)\}_{N=1}^\infty $.
	Assume $ g_N^*(\bx) = \sum_{j=1}^{p_N} g_j(x_j) \in \calG^1 $, $ \sup_N\| g^*_N \|_{\infty} < \infty $, and $ \sup_N\| g^*_N \|_{\ell_0} < \infty $. 
	Suppose that $ K_N \rightarrow \infty $ and $\frac{2^{K_N}\log^2(N)\log(Np_N)}N \rightarrow 0 $ as $N\rightarrow\infty$.
	Granting the noise condition \prettyref{eq:bounded}, regression trees are consistent, that is,
	$$
	\lim_{N \rightarrow\infty}\mathbb{E}_\calD(\|\hat g(T_{K_N}^{\textnormal{\scalebox{0.8}{CART}}})-g^*_N \|^2) = 0.  
	$$
	Furthermore, classification trees are consistent, that is,
	$$
	\lim_{N \rightarrow\infty} (\mathbb{E}_{\calD}({\rm Err}(\hat g(T_{K_N}^{\textnormal{\scalebox{0.8}{C4.5}}}))) - {\rm Err}(g^*_N))
	= 0.
	$$
\end{corollary}
	While \prettyref{cor:consistent_general} allows the component functions of the model to have unbounded variation, it requires the number of relevant variables $\| g^*_N \|_{\ell_0}$ to be uniformly bounded in $N$, in contrast to the $o(\sqrt{K_N})$ growth of $ \|g^*_N\|_{\text{TV}} $ tolerated in \prettyref{cor:consistent}.
	\begin{remark}
		\prettyref{cor:consistent} and \prettyref{cor:consistent_general} do not offer guidance on how to choose the depth $ K_N $. In practice, it is best to let the data decide and therefore cost complexity pruning (i.e., weakest link pruning \citep{breiman1984}) is recommended. This would have one first grow a full tree $ T_{max} $ (to maximum depth) and then minimize
		$$
		\widehat \calR(\hat g(T)) + \lambda |T|
		$$
		over all trees $ T $ that can be obtained from $ T_{max} $ by iteratively merging its internal nodes, where $ \lambda $ is a positive constant and $ |T| $ is the number of terminal nodes of $ T $. Working with the resulting pruned tree enables one to obtain oracle inequalities of the form \prettyref{eq:oracle},
		but with the advantage of having the infimum over both the depth $ K \geq 1 $ and additive functions $ g(\cdot) \in \calG^1 $.
	\end{remark}
	
\subsection{Related consistency results and optimality} \label{sec:consistent_discussion}
Here we compare our consistency results for CART and C4.5 to those of other prediction methods and the optimal minimax rates.

	The reader might be somewhat surprised by the consistency statements in \prettyref{cor:consistent} and \prettyref{cor:consistent_general}, especially since they are qualitatively similar to existing performance guarantees for predictors based on very different principles, like boosting or neural networks.
		For example, \cite[Theorem 1]{buhlmann2006boosting} states that boosting with linear learners is also consistent for a sequence of $ \ell_1 $ constrained linear models $ g_N^*(\bx) = \bbeta^T_N \bx $ on $ [0, 1]^p $ in the high dimensional regime, i.e., when $ \log(p_N) \asymp N^{1-\xi}  $ for $ \xi \in (0, 1) $ and $ \sup_N\|g_N^*\|_{{\rm TV}} = \sup_N\|\bbeta_N\|_{\ell_1} < \infty $.
 
To provide another frame of reference, we also compare the consistency rates of CART \prettyref{eq:rate} and C4.5 \prettyref{eq:rate_logistic} to the corresponding minimax rates for the model classes we consider. 
For regression, the optimal minimax rate (with respect to excess squared error risk) for $s$-sparse additive regression with continuously differentiable component functions is $ \max\{ (s/N)\log(p/s),  sN^{-2/3} \} $ \citep{raskutti2012minimax}, while the optimal rate for additive regression models with bounded total variation $\ell_1$ norm (when $ p \gg N $) is $ \sqrt{(\log(p))/N} $ \citep{tan2019doubly}. Both of these settings, corresponding to $\ell_0 $ and $ \ell_1 $ sparsity, respectively, are covered by our theory for regression trees, and so our $1/\log(N) $ rate \prettyref{eq:rate} is evidently sub-optimal. For classification, the literature is less developed, though there are results for linear logistic regression. For example, \cite{abramovich2019high} show that the optimal minimax rate (with respect to excess mis-classification risk) for $ s$-sparse linear logistic regression is $ \sqrt{(s/N)\log(p/s)} $, again much faster than our $1/\sqrt{\log(N)} $ rate \prettyref{eq:rate_logistic}. 

The sub-optimality of our rates is due to a combination of two factors. First, the form of the decision tree predictions—averaging the response data in the terminal nodes—introduces an inductive bias, which, in the aforementioned $s$-sparse additive regression setting, leads to best-case (yet still sub-optimal) rates of order $ N^{-2/(2+s)} $ \citep{tan2021cautionary}. 
The second source of sub-optimality stems from our method of analysis. Recall from \prettyref{lmm:gain} 
that a key step in our proofs is to lower bound the impurity gain at each node by a constant multiple of the \emph{squared} excess risk, viz.,
$
\mathcal{IG}(\bt) \gtrsim (\widehat \calR_{\bt}(\hat g(T_{K-1})) - \widehat \calR_{\bt}(g^*))^2
$
whenever $ \widehat \calR_{\bt}(\hat g(T_{K-1})) > \widehat \calR_{\bt}(g^*) $.
Had we been able to show (or rather presumed in the form of an assumption) a lower bound
$
\mathcal{IG}(\bt) \gtrsim \widehat \calR_{\bt}(\hat g(T_{K-1})) - \widehat \calR_{\bt}(g^*)
$
whenever $ \widehat \calR_{\bt}(\hat g(T_{K-1})) > \widehat \calR_{\bt}(g^*) $,
then it would have been possible for us to obtain polynomial convergence rates,
instead of logarithmic. In the next section, we obtain a similar lower bound on the impurity gain (\prettyref{lmm:gain_margin}) for separable, large margin data, which does indeed lead to a faster consistency rate.

\section{Beyond discriminative models} \label{sec:margin}

In \prettyref{sec:main}, we
operated under a
discriminative statistical model of the data; that is, we study decision trees under an explicit form of the conditional distribution $\mathbb{P}_{y \mid \bx}$.
In lieu of the logistic regression model from \prettyref{sec:classification}, here we consider
another ubiquitous classification setting in which the data can be perfectly separated into two classes by an additive decision boundary, with margin $\gamma>0$. As we shall see, this setting will allow us to obtain consistency rates which are exponentially faster than those from \prettyref{sec:consistent_rate}.
Separable data assumptions, such as the one formalized in \prettyref{ass:margin} below,
are prevalent in statistical learning literature, especially in the context of (kernel) support vector machines \citep{boser1992training} and boosting \citep{bartlett1998boosting}.
Note that such an assumption is particularly appropriate for our high dimensional setting—when $ p $ is large, there is more freedom for the data to be separated by an additive decision boundary. 
\begin{assumption}[Additively separable, large margin] \label{ass:margin}
	There exists $ \gamma \in (0, 1] $ and $ f^*(\bx) = \sum_{j=1}^{p} f_{j}(x_j) \in \calG^1 $ with $ \max\{\|f^*\|_{{\rm TV}}, \|f^*\|_{\infty}\}\leq 1 $ such that for almost all 
	pairs $(\bx, y)$ drawn from the joint distribution $ \mathbb{P}_{(\bx, y)} $,
	$$
	y f^*(\bx) \geq \gamma.
	$$	
\end{assumption}
\begin{remark}
	If the additive function $ f^*(\bx) = \bbeta^T \bx $ is linear over $[0, 1]^p$, then $\gamma$ corresponds to the maximum
	$\ell_1$-margin subject to $\|\bbeta\|_{\ell_1} \leq 1$, matching the standard margin framework for linearly separable data. 
	\end{remark}

Note that any separable, large margin distributional assumption implies that the Bayes risk is zero; therefore, we aim to show that a tree constructed with C4.5 methodology will have mis-classification risk converging to zero. 
While there are some similarities with the proof of \prettyref{cor:consistent}, the key difference
is an improved lower bound on the impurity gain that establishes nearly linear, rather than quadratic (see \prettyref{lmm:gain}), dependence on the within-node empirical risk.

\begin{lemma}[Information gain for C4.5 with separable, large margin data] \label{lmm:gain_margin}
	Grant \prettyref{ass:margin}.
	Let $K\ge 1$ be any depth and let $\bt$ be any terminal node of $T_{K-1}^{\textnormal{\scalebox{0.8}{C4.5}}}$.
	We have that
	\begin{equation}
	\mathcal{IG}(\bt) \ge \frac{\gamma^2}{30} \cdot \frac{\widehat \calR_\bt(\hat g(T_{K-1}^{\textnormal{\scalebox{0.8}{C4.5}}}))}{\log(1/\widehat \calR_\bt(\hat g(T_{K-1}^{\textnormal{\scalebox{0.8}{C4.5}}})))}.
	\end{equation}
\end{lemma}

Using \prettyref{lmm:gain_margin} in conjunction with \prettyref{eq:training_recursion}
shows that the empirical risk decays sub-exponentially fast in the depth $K$,
much faster than the polynomial
rate of decay for a logistic regression model (\prettyref{thm:training}).
\begin{theorem}[Empirical risk for C4.5 with separable, large margin data] \label{thm:training_margin}
	Granting \prettyref{ass:margin}, for all depths $ K \geq 1 $,
	$$
	\widehat \calR(\hat g(T_K^{\textnormal{\scalebox{0.8}{C4.5}}})) \leq \exp\Big(- \Big(\frac{\gamma^2K}{30} \Big)^{1/2}\Big).
	$$
\end{theorem}
Employing the above empirical risk bound, we can establish the following mis-classification risk bound.

\begin{theorem}[Mis-classification risk for C4.5 with separable, large margin data]\label{thm:oracle_margin}
	Granting \prettyref{ass:margin}, for all depths $ K \geq 1 $,
	\begin{equation*}
	\mathbb{E}_\calD ({\rm Err}(\hat g(T_K^{\textnormal{\scalebox{0.8}{C4.5}}}))) 
	\leq 2 \exp\Big(- \Big(\frac{\gamma^2K}{120} \Big)^{1/2}\Big)+
	C_2\Big(\frac{2^K\log^2(N)\log(Np)}{N}\Big)^{1/4},
	\end{equation*}
	where 
	$C_2$ is the same constant in the statement of \prettyref{thm:oracle}.
\end{theorem}

\prettyref{thm:oracle_margin} immediately implies the following consistency result.
\begin{corollary}[Consistency of C4.5 with separable, large margin data] \label{cor:consistent_margin} 
	Consider a sequence of decision boundaries $\{f^*_N(\cdot)\}_{N=1}^\infty $ with respective margins $\{\gamma_N(\cdot)\}_{N=1}^\infty$
	such that \prettyref{ass:margin} holds.
	Suppose $f^*_N(\bx) = \sum_{j=1}^{p_N} f_{j}(x_j) \in \calG^1 $ and that $K_N \rightarrow\infty$, $ \gamma_N = \omega(1/\sqrt{K_N})  $,
	and $ \frac{2^{K_N}\log^2(N)\log(Np_N)}N \rightarrow 0 $
	as $N \rightarrow\infty$. Then, classification trees are consistent, that is,
	$$
	\lim_{N\to\infty}\mathbb{E}_{\calD} ({\rm Err}(\hat g(T_{K_N}^{\textnormal{\scalebox{0.8}{C4.5}}} ))) = 0.
	$$
\end{corollary}
The hypotheses of \prettyref{cor:consistent_margin} are satisfied if, for example, $ K_N = \lfloor (\xi/2)\log_2(N) \rfloor$, $ \log(p_N) \asymp N^{1-\xi} $ for $ \xi \in (0, 1) $, and $\gamma_N = \gamma $ 
 for some $ \gamma \in (0, 1] $. 
In this case, from \prettyref{thm:oracle_margin}, the consistency rate of C4.5 is 
\begin{equation} \label{eq:rate_margin}
\calO\big(\exp\big(- \sqrt{\gamma^2 \xi \log(N)/240} \big)\big),
\end{equation}
which is sub-polynomial in $N$,
and exponentially faster than the $1/\sqrt{\log(N)}$ rate \prettyref{eq:rate_logistic} for a logistic regression model.

\section{Models with interactions} \label{sec:beyond_additive}
Our main results in \prettyref{sec:main} focused on (ordinary or logistic) additive regression models primarily because notions of approximate and exact sparsity are easier to define and more interpretable in high dimensional settings. However, the interpretation of CART and C4.5 in an (ordinary or logistic) additive regression setting will be muddled by the fact that all interactions it finds will be spurious. It is therefore desirable to have a more comprehensive theory, particularly for data settings where decision trees could be useful.

As we now explain, it is possible to motivate reasonable assumptions
so that our main results for (ordinary or logistic) additive regression models
can be extended to models with interactions.
 To this end, recall the proof outline in \prettyref{sec:main}. We started with the recursion \prettyref{eq:training_recursion} and then used a lower bound on the impurity gain (\prettyref{lmm:gain}), tailored for (ordinary or logistic) additive regression models, to bound the empirical risk (\prettyref{thm:training}) and ultimately show consistency (\prettyref{cor:consistent}). In fact, under suitable assumptions, many of these same ideas work when model class $ \calG $ has interaction terms.
	
	 To illustrate the broad strokes in obtaining these extensions, we will consider the class of $d$-way interaction models
	\begin{align*}
	&  \calG^d \coloneqq \Big\{ g(\bx) \coloneqq \sum_{j_1} g_{j_1}(x_{j_1}) + \sum_{j_1<j_2} g_{j_1,j_2}(x_{j_1},x_{j_2}) 
  \\ & \qquad  \qquad \qquad \qquad\qquad  \qquad \qquad \qquad + \cdots +
	 \sum_{j_1<j_2<\cdots < j_d} g_{j_1,j_2,\dots, j_d}(x_1, x_2, \dots, x_d) \Big\},
	\end{align*}
	which encompasses the additive function class $ \calG^1 $. Thus, models belonging to $ \calG^d $ have interactions involving up to $d$ predictor variables. Conversely, any square-integrable function with interactions involving at most $ d $ variables admits a functional ANOVA decomposition in the form above, where the functional components in the expansion have zero mean and are orthogonal to each other \citep{hooker2007generalized}. We note in passing that, while decision trees can discover interaction effects through the way they are constructed, they do not directly leverage additive structure in the model and so are unlikely to achieve the optimal rates of convergence on $\calG^d$ (this is certainly the case for $ \calG^1 $; see \prettyref{sec:consistent_discussion}). 
		
	By recursing \prettyref{eq:training_recursion}, we can write the empirical risk of the tree as
	\begin{equation} \label{eq:training_recursion2}
	\begin{aligned} 
	\widehat \calR(\hat g(T_K)) &= \widehat \calR(\hat g(T_{K-d})) - \sum_{\bt \in T_{K-d}} \frac{N_\bt}{N}\mathcal{IG}_d(\bt),
	\end{aligned}
	\end{equation}
	where $ \mathcal{IG}_d(\bt) \coloneqq  \mathcal{IG}(\bt) +  \sum_{\bt'} \frac{N_{\bt'}}{N_{\bt}} \mathcal{IG}(\bt') $ is the \emph{$d$-level impurity gain} of a node $\bt \in T_{K-d} $. Here in the definition of $  \mathcal{IG}_d(\bt)$, the sum extends over all descendent nodes $ \bt' $ of $ \bt $ up to depth $ K-1 $. Another way of thinking about $\mathcal{IG}_d(\bt)$ is that it measures the decrease in risk from greedily splitting $d$ times in $\bt$ and thus captures interactions involving up to $ d $ predictor variables. 
For example, according to the representations given by \prettyref{eq:IG} and \prettyref{eq:training_decrease},
 \begin{equation} \label{eq:training_decrease_2}
 \mathcal{IG}(\bt) = \widehat \calR_{\bt}(h(\overline y_{\bt})) -  \min_{\beta_0, \beta_1}\widehat \calR_{\bt}(\beta_0  + \beta_1 \mathbf{1}(x_{j_\bt}> s_\bt)),
\end{equation}
which captures only main effects from splitting \emph{once} in $ \bt$. This explains why \prettyref{lmm:gain} relates the impurity gain to the empirical risk of an (ordinary or logistic) additive regression model. On the other hand, $\mathcal{IG}_2(\bt) = \mathcal{IG}(\bt) + P_{\bt_L} \mathcal{IG}(\bt_L) + P_{\bt_R} \mathcal{IG}(\bt_R)$ equals 
\begin{align*}
&  \widehat \calR_{\bt}(h(\overline y_{\bt})) -  \min_{\beta_0, \beta_1, \beta_2, \beta_3, j_1, s_1, j_2, s_2}\widehat \calR_{\bt}(\beta_0  + \beta_1 \mathbf{1}(x_{j_{\bt}} > s_{\bt}) + \\ & \qquad\qquad\qquad\qquad\qquad\qquad\qquad\qquad  \beta_2 \mathbf{1}(x_{j_1} > s_1, \;x_{j_{\bt}} \leq s_{\bt})+\beta_3 \mathbf{1}(x_{j_2} > s_2, \;x_{j_{\bt}} > s_{\bt})),
\end{align*}
and thus captures both main effects and second order effects from greedily splitting \emph{twice} in $ \bt$. It is then reasonable to postulate that one could relate $ \mathcal{IG}_2(\bt)$ to the empirical risk of a function in $ \calG^2 $. Consequently, a natural generalization of the impurity gain inequality in \prettyref{lmm:gain} to functions in $\calG^d$ would be the following condition.
\begin{assumption} \label{ass:gain_d}
Let $g^*(\cdot) \in \calG^d$ and $ K \geq d $. For any terminal node $ \bt $ of the tree $T_{K-d} $ such that $ \widehat \calR_\bt(\hat g(T_{K-d})) > \widehat \calR_\bt (g^*) $, we have
	\begin{equation} \label{eq:gain4}
		\mathcal{IG}_d(\bt)\geq \frac{(\widehat \calR_\bt(\hat g(T_{K-d})) - \widehat \calR_\bt (g^*))^2}{V^2(g^*)},
	\end{equation}
	for some complexity constant $ V(g^*)$ that depends only on $ g^*(\cdot) $.
\end{assumption}

	Following a similar argument to the one outlined in \prettyref{sec:main}, we can substitute the purported lower bound \prettyref{eq:gain4} into \prettyref{eq:training_recursion2} and use Jensen's inequality to produce the recursion
	$$
	\calE_K \leq \calE_{K-d} \Big(1 - \frac{\calE_{K-d} }{V^2(g^*)} \Big), \quad \calE_{K-d} \geq 0, \quad K \geq d.
	$$
Thus, granting the impurity gain condition \prettyref{eq:gain4}, by \prettyref{lmm:gaind} in \prettyref{supp:non-additive}, 
we have that
\begin{equation} \label{eq:recursion_d}
	\widehat \calR(\hat g(T_K))\leq 
	\widehat \calR(g^*) + 
	\frac{V^2(g^*)d}{K+2d+1}, \quad K \geq d,
\end{equation}
a direct analogue to \prettyref{thm:training}. Thus, the excess empirical risk for a $d$-way interaction model is of order $ d/K $, which means that the depth $ K $ should exceed $ d $ for it to be small. This is to be expected since the depth
$ K $ controls the interaction order of the tree.
Using \prettyref{eq:recursion_d} and
the same steps as the proof of \prettyref{thm:oracle} and \prettyref{cor:consistent}, we can easily deduce that
if $ \{ g^*_N(\cdot) \}_{N=1}^{\infty} $ is a sequence of true models in $ \calG^d $ with $ \sup_N \|g^*_N\|_{\infty} < \infty $ and $ K_N \rightarrow \infty $, $ V(g_N^*) = o(\sqrt{K_N})$, and $\frac{2^{K_N}\log^2(N)\log(Np_N)}N \rightarrow 0$ as $N \rightarrow \infty$, then both regression trees and classification trees are consistent, i.e.,
\begin{equation} \label{eq:consistent_d}
\mathbb{E}_\calD(\| \hat g(T_{K_N}^{\textnormal{\scalebox{0.8}{CART}}}) - g^*_N\|^2) \to 0 \quad \text{and} \quad \mathbb{E}_{\calD}({\rm Err}(\hat g(T_{K_N}^{\textnormal{\scalebox{0.8}{C4.5}}}))) - {\rm Err}(g^*_N) \to 0, \quad \text{as} \quad N \to \infty.
\end{equation}

\section{Random forests} \label{sec:forests}

	The predictive abilities of individual decision trees should intuitively be inherited by random forests due to the ensemble principle and convexity of squared error (see \cite[Proposition 3 and Proposition 4]{denil2014}, \cite[Section 11]{breiman2001random}, or \cite[Section 4.1]{breiman1996bagging}). Indeed, our main results for individual trees in \prettyref{sec:main}
	also carry over to Breiman's random forests \citep{breiman2001random} with relative ease, as we now explain. 
	To keep redundancy to a minimum, we will restrict ourselves to the regression setting  of \prettyref{sec:regression}.
	\subsection{Growing the forest}
	
	Consider a sub-sample $ \calD' $ of size $a_N$ 
	from the original dataset $ \calD $, whereby each sample point is drawn uniformly at random without replacement.\footnote{We deviate slightly from Breiman's original random forests \citep{breiman2001random}, as we do not grow the constituent trees to maximum depth with bootstrapped data. Note, however, that sampling with and without replacement produce similar results when $ a_N = \lfloor N/2 \rfloor$ \citep{friedman2007bagging}. 
	} From this sub-sample, we train a depth $ K $ tree $ T_K^{\textnormal{\scalebox{0.8}{CART}}} $ with CART methodology in the usual way, except that, at each internal node, we select $ m $ (also known as \emph{mtry}) of the $ p $ variables uniformly at random without replacement, as candidates for splitting. That is, for each internal node $ \bt $ of $ T_K^{\textnormal{\scalebox{0.8}{CART}}}  $, we generate a random subset $ \calS \subset \{ 1, 2, \dots, p \} $ of size $ m $ and split along a variable $ x_{\hat \jmath_\bt} $ with split point $ \hat s_{\bt} $, where $(\hat \jmath_\bt, \hat s_{\bt}) \in \argmax_{(j\in\calS,\; s \in \mathbb{R})}\mathcal{IG}(j, s, \bt) $.

We grow $B$ of these depth $K$ regression trees separately using, respectively, $B$ independent realizations $ \bTheta = (\Theta_1, \Theta_2, \dots, \Theta_B)^T $ of a random variable $\Theta$.	 Here $ \Theta$ is distributed according to the law that generates the sub-sampled training data $\calD'$ and candidate variables $\calS$ for splitting at each of the nodes. The output of the $b^{{\rm th}} $ regression tree is denoted by $\hat g(T_K^{\textnormal{\scalebox{0.8}{CART}}} (\Theta_b))$.

	With this notation in place, the random forest output is then simply the empirical average of the $B$ regression tree outputs, namely,
	\begin{equation}\label{eq:random_forest}
	\hat g(\bTheta)(\bx) \coloneqq B^{-1}\sum_{b=1}^B \hat g(T_K^{\textnormal{\scalebox{0.8}{CART}}} (\Theta_b))(\bx).
	\end{equation}
	
	\subsection{Oracle inequality for random forests} \label{sec:oracle_forest}
	By a modification of the proofs of \prettyref{thm:training} and \prettyref{thm:oracle}, it is possible to show the following oracle inequality for random forests. 
	
	\begin{theorem}[Oracle inequality for CART random forests]\label{thm:oracle_forest}
		Granting the noise condition \prettyref{eq:bounded}, for all depths $ K \geq 1 $, we have
		$$
		\mathbb{E}_{\bTheta, \calD}(\| \hat g(\bTheta) - g^*\|^2) \leq 2 \inf_{g(\cdot) \in\calG^1} \Bigg\{  \|g - g^*\|^2 + \frac{p}{m}  \frac{\|g\|^2_{{\rm TV}}}{K+3} + C_1\frac{2^K\log^2(a_N)\log(a_Np)}{a_N} \Bigg\},
		$$
		where $ C_1 $ is the same constant in the statement of \prettyref{thm:oracle}.
		
	\end{theorem}
	\begin{remark} \label{rmk:breiman}
	As the number of trees $ B $ in the forest approaches infinity, 
	by \cite[Theorem 11.1]{breiman2001random},
	$$
	\mathbb{E}_{\bTheta, \calD}(\| \hat g(\bTheta) - g^*\|^2) \rightarrow \mathbb{E}_{\calD}(\| \mathbb{E}_{\bTheta\mid\calD}(\hat g(\bTheta)) - g^*\|^2), \quad \text{almost surely.}
	$$
	Thus, when $ B $ is large, \prettyref{thm:oracle_forest} resembles results such as \prettyref{eq:oracle} in \prettyref{thm:oracle}
	which measure the accuracy of a predictor via the \emph{expected} prediction risk over the data $ \calD $.
	\end{remark}

	To the best of our knowledge, \prettyref{thm:oracle_forest} is one of the first results in the literature that shows explicitly the impact on the prediction risk from randomly choosing subsets of variables as candidates for splitting at the nodes, without any restrictive assumptions on the tree or data generating process. Even though it is not captured by \prettyref{thm:oracle_forest}, empirically, the random variable selection mechanism of forests has the effect of de-correlating and encouraging diversity among the constituent trees, which can greatly improve the performance. It also reduces the computational time of constructing each tree, since the optimal split points do not need to be calculated for every variable at each node. What \prettyref{thm:oracle_forest} does reveal, however, is that this mechanism cannot hurt the prediction risk beyond a benign factor of $ p/m $. In fact, standard implementations of regression forests use a default value of $ m $ equal to $  \lfloor  p/3 \rfloor  $.
	With this choice, we see by comparing \prettyref{thm:oracle_forest} and \prettyref{cor:consistent} that there is essentially no loss in performance (at most a factor of $ p/m = 3 $) over individual trees, despite not optimizing over the full set of variables at the internal nodes. It is also interesting to note that we recover the bound \prettyref{eq:oracle} for individual regression trees when $m = p $.

	\subsection{Consistency of random forests} \label{sec:consistent_forest}
	\prettyref{thm:oracle_forest} immediately implies
	 a consistency result for regression forests, analogous to \prettyref{cor:consistent} for individual regression trees.

	\begin{corollary}[Consistency of random forests] \label{cor:consistent_forest}
		Consider a sequence of prediction problems \prettyref{eq:g^*} with true models $\{g^*_N(\cdot)\}_{N=1}^\infty$.
		Assume that $g^*_N(\bx) = \sum_{j=1}^{p_N} g_{j}(x_j) \in \calG^1$ and $ \sup_N \|g^*_N\|_{\infty} < \infty $.
		Suppose that $ K_N \rightarrow \infty$, 
		$ \|g_N^*\|_{{\rm TV}} = o(\sqrt{(m_N/p_N)K_N}) $, and $\frac{2^{K_N}\log^2(a_N)\log(a_Np_N)}{a_N} \rightarrow 0 $ as $ N \rightarrow \infty$.
		Then, granting the noise condition \prettyref{eq:bounded}, regression forests 
		are consistent, that is,
		$$ \lim_{N\rightarrow\infty}\mathbb{E}_{\bTheta, \calD}(\| \hat g(\bTheta)-g^*\|^2) = 0. $$
	\end{corollary}

	 Like the consistency statement for CART in \prettyref{cor:consistent},
	  the hypotheses of \prettyref{cor:consistent_forest} are satisfied if, for example, $ K_N = \lfloor (\xi/2)\log_2(a_N) \rfloor$, $ \log(p_N) \asymp a_N^{1-\xi}$, and $ m_N = \lfloor p_N/3 \rfloor$, for some constant $ \xi \in (0, 1) $, yielding the same $\calO(1/\log(N))$ rate as a single tree (see \prettyref{eq:rate}).
	  It is also interesting to note that consistency is still possible even if only a vanishing fraction of variables are randomly selected at each node, i.e., $m_N = o(p_N)$.
	At the extreme end, consistency holds even when $ m_N\equiv 1 $; that is, only a single coordinate is selected at random at each node, provided appropriate restrictions are placed on $ p_N $ and $ \|g^*_N\|_{\text{TV}}$.

	\prettyref{cor:consistent_forest} provides a partial answer to a problem posed by \cite{scornet2015consistency}: 
	\begin{quotation}
		\noindent\normalsize\textit{It remains that a substantial research effort is still needed to understand the properties of forests in a high-dimensional setting, when $p = p_N$ may be substantially larger than the sample size. }
	\end{quotation}
	More specifically, \prettyref{cor:consistent_forest} strengthens \cite[Theorem 1]{scornet2015consistency}, which shows that random forests are consistent for additive regression when $ p $ is fixed, the component functions are continuous, and $ K_N \rightarrow \infty $ and $\frac{2^{K_N}(\log^9(a_N))}{a_N} \rightarrow 0 $ as $ N \rightarrow \infty $.
	In contrast, here we allow the dimensionality $p_N$ to grow sub-exponentially with the sample size $N$ (under $ \ell_1 $ sparsity constraints) and also for the component functions to be possibly discontinuous. In fact, the proof of \prettyref{cor:consistent_general} can be modified to establish consistency of random forests for additive regression with growing dimensionality $p_N$ (under $ \ell_0 $ sparsity constraints) and Borel measurable component functions—without assuming continuity or finite total variation. 
	More specifically, suppose that $ \sup_N\| g^*_N \|_{\infty} < \infty $, $ \sup_N\| g^*_N \|_{\ell_0} < \infty $, and $ m_N \asymp p_N $.
	If $ K_N \rightarrow \infty $
	and $\frac{2^{K_N}\log^2(a_N)\log(a_Np_N)}{a_N} \rightarrow 0 $ as $N\rightarrow\infty$, then granting the noise condition \prettyref{eq:bounded}, we have
	$$
	\lim_{N \rightarrow\infty}\mathbb{E}_{\bTheta,\calD}(\|\hat g(\bTheta)-g^*_N \|^2) = 0.
	$$

We close this section by saying that it is still largely a mystery (at least theoretically) why bagging and the random feature selection mechanism are so effective at reducing the prediction risk. Our bounds in  \prettyref{thm:oracle_forest} only show that these apparatuses do not degrade the performance beyond small factors. Certainly more work needs to be done to answer these questions.

\subsection{Empirical studies} \label{sec:empirical}

Because random forests are now a classic topic in machine learning and data science, they have been subject to thorough empirical scrutiny and investigation under a variety of model specifications. This includes the high dimensional regime when $ p \gg N $. As the literature is vast, we only mention a few experimental papers below.

Regarding the $m$ parameter, \cite{genuer2008random} provide a detailed empirical analysis of random forests in the high dimensional setting. They consider both synthetic and real-world data for regression and classification tasks. Among the simulated examples for regression are Friedman's benchmark models \citep{breiman1996bagging}, which, in our notation, belong to the classes $ \calG^2 $ and $ \calG^4 $ and involve either $4$ or $5$ relevant predictor variables. For a sample size of $ N = 100 $, the simulation results (see Figures 3-5) reveal that forests are quite robust to the inclusion of noisy predictor variables (with ambient dimensionality ranging from $ p = 100 $ to $ p = 1000 $). The author finds that forests achieve best performance when $m$ is set to be equal to the ambient dimension $ p $, corresponding to bagging. Similar observations are made in \citep[Table I]{segal2004machine} for one of Friedman's models with $N=200$ and $p=510$.

An investigation of the tree depth in random forests can be found in \citep[Figures 10 \& 11]{zhou2021trees}, where they sample $ N=50 $ and $ N=100$ data points from a sparse linear model with $5$ and $10$ relevant variables, and ambient dimensionality $ p = 1000 $. Shallow trees are found to be advantageous when the model has a low signal-to-noise ratio. In \citep{duroux2018impact}, Figures A.1 \& A.3 reveal that forests with smaller trees achieve similar performance to forests with fully grown trees if the number of terminal nodes is properly tuned. One of their examples (Model 8) consists of a sparse additive model with $4$ relevant variables, $ p = 1000 $, and $ N = 500 $.

Other references on the impact of hyperparameter tuning in random forests can be found in the review article \citep{probst2019hyper}.

It is still largely a mystery (at least theoretically) why bagging and the random feature selection mechanism are so effective at reducing the prediction risk. Our bounds in  \prettyref{thm:oracle_forest} only show that these apparatuses do not degrade the performance beyond small factors. Fascinating recent work by \cite{mentch2020randomization} shows that $ m $ plays a similar role as the shrinkage penalty in explicitly regularized procedures. More specifically, when $ p > N $, they show that if an ensemble predictor is formed by averaging over many linear regression models with orthogonal designs and randomly selected subsets of variables, then asymptotically as the number of models goes to infinity, the coefficient vector of the ensemble is shrunk by a factor of $ m/p $. It is possible that a similar form of implicit regularization is occurring in our high dimensional additive setting, which may lead to even better performance over individual trees. Certainly more work needs to be done to answer these questions.

\section{Conclusion} \label{sec:discussion}

In this paper, we showed that decision trees and random forests adapt to $\ell_0$ or $\ell_1$ forms of sparsity and can accommodate (essentially) arbitrary types of predictor variables, such as continuous, discrete, and/or dependent.
 Our work is primarily of theoretical value, since we study existing decision tree methodology, namely, CART and C4.5. Nevertheless, given their widespread popularity in many applied disciplines, we believe our results can be used as a theoretical justification for practical application on high dimensional regression and classification problems. Specifically, \prettyref{thm:oracle} and \prettyref{thm:oracle_forest} give an explicit characterization of how the various quantities (e.g., tree depth, mtry, sample size, ambient dimension, sparsity level) interact with each other and determine the performance, thereby motivating specific choices of the parameters (e.g., tree depth, mtry). Furthermore, consistency in \prettyref{cor:consistent} and \prettyref{cor:consistent_forest} serves as a \emph{stress-test} and shows how practical implementations of decision trees and random forests can be accurate even with a very high predictor variable count.

\appendix
\numberwithin{equation}{section}
\bigskip
\begin{center}
{\large\bf SUPPLEMENTARY MATERIAL}
\end{center}

In this supplement, we provide the complete details of the proofs of all of our results. 
Apart from a section where we introduce some preliminaries,
the organization of this supplement follows the organization of the main body.

\section{Preliminaries} \label{supp:prelim}

In this section, we introduce some key ideas and notation which will be useful when analyzing CART and C4.5 separately.

\subsection{CART Preliminaries} \label{supp:CART} 

For multivariate functions $ u(\cdot) $ and $ v(\cdot) $, let
\begin{equation} \label{eq:emp_norm}
\|u\|^2_{\calD}\coloneqq \frac{1}{N} \sum_{i=1}^N (u(\bx_i))^2 \quad \text{and} \quad \langle u,  v\rangle_{\calD} \coloneqq \frac{1}{N}\sum_{i=1}^N u(\bx_i)v(\bx_i)
\end{equation}
be the squared norm and inner product, respectively, with respect to the empirical measure on the data. 
Similarly, we
define the within-node squared norm and inner product, respectively, as
\begin{equation} \label{eq:emp_normt}
\|u\|^2_\bt \coloneqq \frac{1}{N_\bt} \sum_{\bx_i \in \bt} (u(\bx_i))^2 \quad \text{and} \quad \langle u,  v\rangle_\bt \coloneqq \frac{1}{N_\bt}\sum_{\bx_i\in \bt} u(\bx_i)v(\bx_i).
\end{equation}
We will also extend our notation to the response data and write, for example, $ \|y-u\|^2_{\bt} = \frac{1}{N_{\bt}}\sum_{\bx_i \in \bt} (y_i-u(\bx_i))^2 $, $ \langle y, u \rangle_{\bt} = \frac{1}{N_{\bt}}\sum_{\bx_i \in \bt} y_i u(\bx_i)  $, $ \|y-u\|^2_{\calD}= \frac{1}{N}\sum_{i=1}^N (y_i-u(\bx_i))^2 $, and $ \langle y, u \rangle_{\calD} = \frac{1}{N}\sum_{i=1}^N y_i u(\bx_i)  $.

Define the standardized decision stump corresponding to splitting a node $\bt$ at variable $x_j$ with splitting value $s$ by 
\begin{equation} \label{eq:stumps} 
\Psi_{\bt}(\bx)   \coloneqq  
\beta_0+\beta_1\mathbf{1}(x_{j} > s)\in \calG^1, \quad \beta_0 \coloneqq \frac{N_{\bt_R}}{\sqrt{N_{\bt_L}N_{\bt_R}}}, \quad \beta_1 \coloneqq - \frac{N}{\sqrt{N_{\bt_L}N_{\bt_R}}},
\end{equation} 
where we have omitted the dependence of $\Psi_{\bt}(\cdot)$, $\beta_0$, $\beta_1$, $\bt_L = \{\bx \in \bt: x_j \le s\}$ and $\bt_R = \{\bx \in \bt: x_j > s\}$ on $(j,s)$ for notational convenience. 	From \prettyref{eq:stumps} we have that 
\begin{equation} \label{eq:stumps2}
\Psi_\bt(\bx) \mathbf{1}(\bx \in \bt) = \frac{ \mathbf{1}(\bx \in \bt_L) N_{\bt_R} - \mathbf{1}(\bx \in \bt_R) N_{\bt_L} }{\sqrt{N_{\bt_L} N_{\bt_R}} }.
\end{equation}

Recall that $\calL(\cdot, \cdot)$ is the squared error loss \prettyref{eq:training_loss_r} and, as such, the impurity gain \prettyref{eq:impurity_r} corresponds to the reduction in variance.
\prettyref{lmm:hilbert} shows that this impurity gain can be written as the squared within-node correlation between the residuals and a standardized decision stump.
\begin{lemma}[Information gain as an inner product] \label{lmm:hilbert}
	Let $\bt$ be any node in $T^{\textnormal{\scalebox{0.8}{CART}}}$. We have
	\begin{equation*} \label{eq:cor_iden}
	\mathcal{IG}(j,s,\bt) = |\langle  y - \overline y_{\bt}, \Psi_{\bt} \rangle_{\bt}|^2.
	\end{equation*}
\end{lemma}
\begin{proof}
	By \prettyref{eq:stumps2}, we have that
	\begin{align*}
	|\langle  y - \overline y_{\bt}, \Psi_{\bt} \rangle_{\bt} |^2 
	= \Big|\frac{ (N_{\bt_L}\overline y_{\bt_L}) N_{\bt_R} - (N_{\bt_R} \overline y_{\bt_R}) N_{\bt_L} }{N_{\bt}\sqrt{N_{\bt_L}N_{\bt_R}} } \Big|^2
	= \frac{N_{\bt_L}N_{\bt_R}}{N^2_{\bt}} (\overline y_{\bt_L} - \overline y_{\bt_R})^2
	= \mathcal{IG}(j,s,\bt),
	\end{align*}
	where the final equality comes from \cite[Section 9.3]{breiman1984}.
\end{proof}

\subsection{C4.5 Preliminaries} \label{supp:C4.5} 
For convenience we will use the notation $\beta_\bt$ to denote the within-node output of a classification tree $T_K^{\textnormal{\scalebox{0.8}{C4.5}}}$, i.e.,
\begin{equation} \label{eq:beta_t}
\beta_{\bt} \coloneqq  \log\Big( \frac{\eta_\bt}{1-\eta_\bt} \Big) = \hat g(T_K^{\textnormal{\scalebox{0.8}{C4.5}}})(\bx), \quad \bx \in \bt,
\end{equation}
where the second equality comes from
\prettyref{eq:h_c}.
Recall that the within-node empirical risk of an arbitrary C4.5 tree $\widehat\calR_{\bt}( \hat g(T_K^{\textnormal{\scalebox{0.8}{C4.5}}}))$ corresponds to the within-node entropy \prettyref{eq:impurity_c}, i.e.,
\begin{equation} \label{eq:log2_t}
\widehat\calR_{\bt}( \hat g(T_{K}^{\textnormal{\scalebox{0.8}{C4.5}}})) = \eta_{\bt} \log(1/\eta_{\bt}) + (1 - \eta_{\bt}) \log(1/(1-\eta_\bt)) \leq \log(2).
\end{equation}
It follows that the empirical risk is similarly bounded:
\begin{equation}\label{eq:log2_N}
\widehat\calR( \hat g(T_{K}^{\textnormal{\scalebox{0.8}{C4.5}}})) = \sum_{\bt \in T_K}\frac{N_{\bt}}{N} \widehat\calR_{\bt}( \hat g(T_{K}^{\textnormal{\scalebox{0.8}{C4.5}}}))
\leq \log(2).
\end{equation}
Though we do not measure the out of sample performance with the population (also known as the logistic) risk for C4.5, it will be a useful tool in our proofs. Recall that the population risk \prettyref{eq:pop_loss} for C4.5 is
\begin{equation} \label{eq:test_loss}
\calR(g) =  \mathbb{E}_{(\bx,y)} \log(1+\exp(-y g(\bx))) = \mathbb{E}_{(\bx,y)} q(y g(\bx)),
\end{equation}
where we define $ q(z) \coloneqq \log(1+\exp(-z)) $ for convenience. We will frequently use the first and second derivatives
\begin{equation} \label{eq:q' and q''}
q'(z) = \frac{-e^{-z}}{1+e^{-z}}, \quad q''(c) = \frac{e^{-z}}{(1+e^{-z})^2} = \frac{e^{-|z|}}{(1+e^{-|z|})^2} \le \frac{1}{4}
\end{equation}
to approximate $q(\cdot)$ with a second order Taylor expansion:
\begin{equation}\label{eq:taylor}
(b-a)q'(a) \leq q(b) - q(a) \leq (b-a)q'(a) +
q''(c)(b-a)^2,
\end{equation}
where $ c $ is some number between $ \min\{a, b\} $ and $ \max\{a, b\} $.

\section{Main results}

We provide the complete proofs of the results in \prettyref{sec:main}, namely, \prettyref{lmm:gain} for the impurity gain lower bound, \prettyref{thm:training} for the empirical risk bound, and \prettyref{thm:oracle} for the oracle inequality.

\subsection{Impurity gain}\label{supp:gain}

We separately prove \prettyref{lmm:gain} for CART and C4.5.

\subsubsection{Impurity gain for CART} \label{supp:gain_CART}

Before we prove \prettyref{lmm:gain} for CART, we first introduce some key ideas.

We use recent tools for the analysis of decision trees from \cite{klusowski2020sparse}, namely, Lemma A.4 in the supplement therein. Let $g(\cdot)$ be any additive function in $\calG^1$.
Define an empirical measure $ \Pi(j, s) $ on split points $ s $ and variables $ x_j $, having Radon-Nikodym derivative (with respect to Lebesgue measure and counting measure)
\begin{equation} \label{eq:prior_cart}
\frac{d\Pi(j, s)}{d(j, s)} \coloneqq 
\frac{|D g_j (s)|\sqrt{P_{\bt_L}P_{\bt_R}}}{\sum_{j'=1}^p \int |D g_{j'} (s')|\sqrt{P_{\bt'_L}P_{\bt'_R}}ds'},
\end{equation}
where $(\bt_L, \bt_R)$ and $(\bt'_L, \bt'_R)$ are the daughter nodes corresponding to the splits $(j,s)$ and $(j',s')$, respectively. We use the shorthand $ D g_j (\cdot) $ to denote the divided difference of $ g_j(\cdot) $ for the successive ordered data points along the $j^{\text{th}}$ direction within $ \bt $. 
That is, if $ x_{(1)j} \leq x_{(2)j} \leq \cdots \leq x_{(N_{\bt})j} $ denotes the ordered data along the $j^{\text{th}}$ direction within $ \bt $, then 
\begin{equation} \label{eq:D}
D g_j (s) \coloneqq 
\begin{cases}
\frac{g_j(x_{(i+1)j})-g_j(x_{(i)j})}{x_{(i+1)j}-x_{(i)j}} & \text{if} \; x_{(i)j} \leq s < x_{(i+1)j} \\
0 & \text{if} \; s= x_{(i)j} = x_{(i+1)j} , \; s<x_{(1)j}, \; \text{or} \; s>x_{(N_{\bt})j} \\
\end{cases}.
\end{equation}
Observe that $D g_j (\cdot) $ coincides with the derivative of the function that linearly interpolates the points $ \{(x_{(1)j},g_j(x_{(1)j})), (x_{(2)j}, g_j(x_{(2)j})), \dots, (x_{(N_{\bt})j}, g_j(x_{(N_{\bt})j})) \} $.

We are now ready to prove \prettyref{lmm:gain} for CART.

\begin{proof}[Proof of \prettyref{lmm:gain} for CART]
Since $ \mathcal{IG}(\bt) $ is by definition the maximum of $ \mathcal{IG}( j, s, \bt) $ over $j$ and $s$, we have from the fact that a maximum is larger than average
\begin{equation} \label{eq:main_lower}
\mathcal{IG}(\bt) 
\geq \int \mathcal{IG}(j, s, \bt)d\Pi(j, s)
= \int |\langle  y - \overline y_{\bt}, \Psi_{\bt} \rangle_{\bt} |^2d\Pi(j, s),
\end{equation}
where the last identity follows from
\prettyref{lmm:hilbert}.
Next, by Jensen's inequality for the square function, \prettyref{eq:main_lower} is further lower bounded by
\begin{equation} \label{eq:del_lower}
\int |\langle  y - \overline y_{\bt}, \Psi_{\bt} \rangle_{\bt} |^2d\Pi(j, s)  \geq \Big( \int |\langle  y - \overline y_{\bt}, \Psi_{\bt} \rangle_{\bt}| d\Pi(j, s) \Big)^2.  
\end{equation}
We now evaluate the expression in the right hand side of \prettyref{eq:del_lower}.
Rewriting \prettyref{eq:stumps2} as
$$
\sqrt{P_{\bt_L}P_{\bt_R}}\Psi_{\bt}\mathbf{1}(\bx \in \bt) =  \mathbf{1}(\bx \in \bt_L) P_{\bt_R} - \mathbf{1}(\bx \in \bt_R)P_{\bt_L} = -(\mathbf{1}(x_j > s)-P_{\bt_R})\mathbf{1}(\bx\in\bt),
$$ 
we see that $ \sqrt{P_{\bt_L}P_{\bt_R}}\langle  y - \overline y_{\bt}, \Psi_{\bt} \rangle_{\bt} =  -\langle  y - \overline y_{\bt}, \mathbf{1}(x_j > s) \rangle_{\bt} $, which implies that
\begin{equation} \label{eq:lower_abs}
\int |\langle  y - \overline y_{\bt}, \Psi_{\bt} \rangle_{\bt}| d\Pi(j, s) = \frac{\sum_{j=1}^p \int  |D g_j (s)||\langle  y - \overline y_{\bt}, \mathbf{1}(x_j > s) \rangle_{\bt}|ds }{\sum_{j'=1}^p \int |D g_{j'} (s')|\sqrt{P_{\bt'_L}P_{\bt'_R}}ds'}.
\end{equation}

Continuing on the numerator in \prettyref{eq:lower_abs}, we use the fact that the integral of the absolute value is at least the absolute value of the integral, yielding
\begin{equation} \label{eq:lower_Num}
\sum_{j=1}^p \int  |D g_j (s)||\langle  y - \overline y_{\bt}, \mathbf{1}(x_j > s) \rangle_{\bt}|ds \geq \Big | \sum_{j=1}^p \int  D g_j (s)\langle  y - \overline y_{\bt}, \mathbf{1}(x_j > s) \rangle_{\bt}ds \Big|.
\end{equation}
Using linearity of the inner product and integration and the fundamental theorem of calculus, the expression in the right hand side of \prettyref{eq:lower_Num} can be simplified as follows
\begin{equation} \label{eq:lower_int}
\begin{aligned}
\sum_{j=1}^p \int  D g_j (s)\langle  y - \overline y_{\bt}, \mathbf{1}(x_j > s) \rangle_{\bt}ds & = \Big\langle  y - \overline y_{\bt}, \sum_{j=1}^p \int D g_j (s)\mathbf{1}(x_j > s)ds \Big\rangle_{\bt} \\ & = 
\Big\langle  y - \overline y_{\bt}, \sum_{j=1}^p g_j\Big \rangle_{\bt} \\ & = \langle  y-\overline y_{\bt},  g\rangle_{\bt}.
\end{aligned}
\end{equation}
Combining
\prettyref{eq:main_lower}, \prettyref{eq:del_lower}, \prettyref{eq:lower_abs}, \prettyref{eq:lower_Num}, and
\prettyref{eq:lower_int}
 proves that
\begin{equation} \label{eq:lower_pen}
\mathcal{IG}(\bt) \geq \frac{|\langle  y-\overline y_{\bt},  g\rangle_{\bt}|^2}{\big(\sum_{j'=1}^p \int |D g_{j'} (s')|\sqrt{P_{\bt'_L}P_{\bt'_R}}ds'\big)^2}.
\end{equation}
Our next goal is to provide respective lower and upper bounds on the numerator and denominator of \prettyref{eq:lower_pen}.
For the denominator of \prettyref{eq:lower_pen}, note that for each $ j' $,
\begin{align*}
\int |D g_{j'} (s')|\sqrt{P_{\bt'_L}P_{\bt'_R}}ds'
& =  \sum_{i=1}^{{N_{\bt}}-1}\int_{x_{(i)j'}}^{x_{(i+1)j'}} |D g_{j'} (s')|ds'\sqrt{(i/{N_{\bt}})(1-i/{N_{\bt}})} \\
& = \sum_{i=1}^{{N_{\bt}}-1}|g_{j'}(x_{(i+1)j'})-g_{j'}(x_{(i)j'})|\sqrt{(i/{N_{\bt}})(1-i/{N_{\bt}})} \\
& \leq \frac{{\rm TV}(g_{j'})}{2},
\end{align*}
and hence, summing over $ j' = 1, 2, \dots, p $, we obtain
\begin{equation} \label{eq:denom}
\sum_{j'=1}^p \int |D g_{j'} (s')|\sqrt{P_{\bt'_L}P_{\bt'_R}}ds' \leq \frac{\|g\|_{\rm TV}}{2}.
\end{equation}

For the numerator of \prettyref{eq:lower_pen}, we use the Cauchy-Schwarz inequality to lower bound $ \langle  y-\overline y_{\bt},  g\rangle_{\bt} $ by
$$
\langle  y-\overline y_{\bt},  g\rangle_{\bt} = \langle  y-\overline y_{\bt},  y \rangle_{\bt} + \langle  y-\overline y_{\bt}, g- y \rangle_{\bt} \geq \| y-\overline y_{\bt}\|^2_{\bt} - \| y-\overline y_{\bt}\|_{\bt} \|  y-g\|_{\bt},
$$
where we also used $ \langle  y-\overline y_{\bt},  y \rangle_{\bt} = \| y-\overline y_{\bt}\|^2_{\bt} $.
Now, by the AM–GM inequality, $  \| y-\overline y_{\bt}\|_{\bt} \|  y-g\|_{\bt}  \leq ( \| y-\overline y_{\bt}\|^2_{\bt} + \|  y-g\|^2_{\bt})/2 $, and hence $ \langle  y-\overline y_{\bt},  g\rangle_{\bt} \geq (\|  y - \overline y_{\bt}\|^2_{\bt} - \|  y-g\|^2_{\bt})/2$.
 Squaring both sides and using the assumption $ \|  y - \overline y_{\bt}\|^2_{\bt} - \|  y-g\|^2_{\bt}\geq 0 $, we have
\begin{equation} \label{eq:num}
|\langle  y-\overline y_{\bt},  g\rangle_{\bt}|^2 \geq \frac{(\|  y - \overline y_{\bt}\|^2_{\bt} - \|  y-g\|^2_{\bt})^2}{4}.
\end{equation}
Applying inequalities \prettyref{eq:denom} and \prettyref{eq:num} to \prettyref{eq:lower_pen}, we therefore have shown that
$$
\mathcal{IG}(\bt) \geq 
\frac{(\|  y - \overline y_{\bt}\|^2_{\bt} - \|  y-g\|^2_{\bt})^2}{\|g\|^2_{\rm TV}} = \frac{(\widehat\calR_{\bt}( \hat g(T_{K-1})) - \widehat\calR_{\bt}( g))^2}{\|g\|_{\rm TV}^2},
$$
which completes the proof.
\end{proof}

\subsubsection{Information gain for C4.5} \label{supp:gain_id3}

The overall structure of the proof of \prettyref{lmm:gain} for C4.5
is similar to the structure of the proof of \prettyref{lmm:gain} for CART in 
\prettyref{supp:gain_CART}.

Let $g(\cdot)$ be any additive function in $\calG^1$. Like \prettyref{eq:prior_cart}, we introduce a similar empirical measure $\Pi(j, s)$ on the variables and split points:
\begin{equation} \label{eq:prior_id3}
\frac{d\Pi(j, s)}{d(j,s)} \coloneqq \frac{|D g_j (s)|}{\sum_{j'=1}^p \int |Dg_{j'}(s')|ds'}.
\end{equation}
Using the definitions \prettyref{eq:D} and \prettyref{eq:prior_id3}, we have
\begin{equation} \label{eq:prior}
\int \text{sgn}(D g_j (s))\mathbf{1}(x_{ij} > s) d\Pi(j, s) = \sum_{j=1}^p\frac{g_j(x_{ij})-g_j(x_{(1)j})}{\sum_{j'=1}^p \int |Dg_{j'}(s')|ds'} = \frac{g(\bx_{i})- g(\bx_{(1)})}{\sum_{j'=1}^p \int |Dg_{j'}(s')|ds'}
\end{equation}
and
\begin{equation} \label{eq:TV}
\sum_{j'=1}^p \int |Dg_{j'}(s')|ds' = \sum_{j'=1}^p\sum_{i=1}^{N_{\bt}-1}|g_{j'}(x_{(i+1)j'})-g_{j'}(x_{(i)j'})| \leq \|g\|_{\rm TV}.
\end{equation}

Before proving \prettyref{lmm:gain} for C4.5, we will prove the following statement.
\begin{lemma} \label{lmm:gainv}
	For any $ v\in [0, 1] $ and any additive $g(\cdot) \in \calG^1$, 
	\begin{equation} \label{eq:IG-lower3}
\mathcal{IG}( \bt) 
\geq v (\widehat \calR_{\bt}( \hat g(T_{K-1}^{\textnormal{\scalebox{0.8}{C4.5}}})) - \widehat \calR_{\bt}( g) )
-v^2\Big(\frac{(\|g\|_{\rm TV}+ \|g\|_{\infty})^2}{4}+\frac{\exp(v(\|g\|_{\rm TV}+ \|g\|_{\infty}))}{(1-v)^2}\Big).
\end{equation}
\end{lemma}
\begin{proof}
	Recalling \prettyref{eq:beta_t},
	 we define the decision stump
	\begin{equation}\label{eq:beta_stump}
	\begin{aligned}
	& \beta_{\bt}(i,j, s) = \beta_{\bt}(1-v) + \beta(g) \mathbf{1}(x_{ij}>s),\\
	& \beta(g) \coloneqq v \cdot 
	\Big(
	\text{sgn}(D g_j (s))\cdot\sum_{j'=1}^p \int |Dg_{j'}(s')|ds' 
	+ g(\bx_{(1)})\Big).
	\end{aligned}
	\end{equation}
	By \prettyref{eq:TV} 
	we have that
	\begin{equation} \label{eq:beta2}
	|\beta(g)| \le v \Bigg(\sum_{j'=1}^p \int |Dg_{j'}(s')|ds'+ |g(\bx_{(1)})| \Bigg) \leq v(\|g\|_{\rm TV}+ \|g\|_{\infty}).
	\end{equation}
	From \prettyref{eq:training_decrease_2}
	we have
	\begin{equation} \label{eq:IG-lower}
	\begin{aligned}
	\mathcal{IG}( \bt)  &= 
	 \widehat \calR_\bt(\beta_\bt)- \min_{\beta_0,\beta_1,j,s} \widehat\calR_\bt(\beta_0+\beta_1 \mathbf{1}(x_j>s)) \\
	&\geq \frac{1}{N_{\bt}}\sum_{\bx_i \in \bt} \int (q(y_i \beta_{\bt}) - q(y_i\beta_{\bt}(i,j, s)))d\Pi(j, s).
	\end{aligned}
	\end{equation}
	Applying \prettyref{eq:taylor} to \prettyref{eq:IG-lower}, we see that
	\begin{equation} \label{eq:taylor2}
	\begin{aligned}
	&\frac{1}{N_{\bt}}\sum_{\bx_i \in \bt} \int (q(y_i \beta_{\bt}) - q(y_i\beta_{\bt}(i,j, s)))d\Pi(j, s) \\
	& \geq \frac{1}{N_{\bt}} \sum_{\bx_i \in \bt} \Big( \int(y_i \beta_{\bt} - y_i\beta_{\bt}(i,j, s)) q'(y_i \beta_{\bt})d\Pi(j, s) 
	- \frac{v^2(|\beta_{\bt}|+\|g\|_{\rm TV} +  \|g\|_{\infty})^2}{2} q''(c_i) \Big),
	\end{aligned}
	\end{equation}
	where $c_i$ is between $y_i \beta_{\bt}$ and $y_i\beta_{\bt}(i,j, s)$ and we used the fact that $|y_i\beta_{\bt} - y_i\beta_{\bt}(i,j, s)|  \le v(|\beta_\bt|+\|g\|_{\rm TV}+ \|g\|_{\infty})$ from \prettyref{eq:beta2}.
	
	We first bound the second term in \prettyref{eq:taylor2}. From \prettyref{eq:q' and q''} and \prettyref{eq:beta2} we have that
	$$
	q''(c_i)
	 \leq \exp(-|c_i|) \leq \exp(-(1-v)|\beta_{\bt}|+v(\|g\|_{\rm TV}+\|g\|_{\infty}))
	$$
	so that
	\begin{equation} \label{eq:beta^2}
	\beta_{\bt}^2 q''(c_i) \leq \beta^2_{\bt}\exp(-(1-v)|\beta_{\bt}|+v(\|g\|_{\rm TV}+\|g\|_{\infty}) ) 
	\leq 
	\frac{\exp(v(\|g\|_{\rm TV}+\|g\|_{\infty}))}{(1-v)^2},
	\end{equation}
	where we used $x^2\exp(-(1-v)x) \le 1/(1-v)^2$.
	Combining \prettyref{eq:beta^2} with $q''(c_i) \le 1/4$ from \prettyref{eq:q' and q''}, it follows that
	\begin{equation} \label{eq:second}
	\begin{aligned}
	\frac{v^2(|\beta_{\bt}|+\|g\|_{\rm TV}+\|g\|_{\infty})^2}{2} q''(c_i) & \leq v^2(\beta^2_{\bt}+(\|g\|_{\rm TV}+\|g\|_{\infty})^2) q''(c_i)  \\ 
	& \leq v^2\Big(\frac{(\|g\|_{\rm TV}+\|g\|_{\infty})^2}{4}+\frac{\exp(v(\|g\|_{\rm TV}+\|g\|_{\infty}))}{(1-v)^2}\Big).
	\end{aligned}
	\end{equation}
	Now we aim to  bound the first term in \prettyref{eq:taylor2}.
	Substituting \prettyref{eq:beta_stump} and using the identity \prettyref{eq:prior}, we have that 
	\begin{equation} \label{eq:first}
	\begin{aligned}
	 & \frac{1}{N_{\bt}}\sum_{\bx_i \in \bt} \int(y_i \beta_{\bt} - y_i\beta_{\bt}(i,j, s)) q'(y_i \beta_{\bt})d\Pi(j, s) \\
	& \qquad\qquad= \frac{v}{N_{\bt}}\sum_{\bx_i \in \bt} \int y_i (\beta_\bt + \beta(g) \mathbf{1}(x_{ij} >s)) q'(y_i \beta_{\bt})d\Pi(j, s) \\
	& \qquad\qquad= \frac{v}{N_{\bt}}\sum_{\bx_i \in \bt} (y_i \beta_{\bt} -y_i g(\bx_i)) q'(y_i \beta_{\bt})\\
	\end{aligned}
	\end{equation}
	Using \prettyref{eq:taylor} we can lower bound \prettyref{eq:first}:
	\begin{equation} \label{eq:first2}
	\begin{aligned}
	\frac{v}{N_{\bt}}\sum_{\bx_i \in \bt} (y_i \beta_{\bt} -y_i g(\bx_i)) q'(y_i \beta_{\bt}) 
	  &\ge  \frac{v}{N_{\bt}}\sum_{\bx_i \in \bt}(q(y_i \beta_{\bt}) -q(y_i g(\bx_i))) \\
	&= v (\widehat \calR_{\bt}( \hat g(T_{K-1}^{\textnormal{\scalebox{0.8}{C4.5}}})) - \widehat \calR_{\bt}( g) ).
	\end{aligned}
	\end{equation}
	 Combining \prettyref{eq:IG-lower}, \prettyref{eq:taylor2},  \prettyref{eq:second}, \prettyref{eq:first}, and \prettyref{eq:first2} completes the proof.
\end{proof}

We are now ready to complete the proof of \prettyref{lmm:gain} for C4.5.

\begin{proof}[Proof of \prettyref{lmm:gain} for C4.5]
	In \prettyref{lmm:gainv}, choose 
	\begin{equation}\label{eq:v}
	v =  \frac{2(\widehat\calR_{\bt}( \hat g(T_{K-1}^{\textnormal{\scalebox{0.8}{C4.5}}})) - \widehat\calR_{\bt}( g)) }
	{(\|g\|_{\rm TV}+\|g\|_{\infty})^2+8},
	\end{equation}
	which is non-negative by the assumption that $ \widehat\calR_{\bt}( \hat g(T_{K-1}^{\textnormal{\scalebox{0.8}{C4.5}}})) > \widehat\calR_{\bt}(g) $.
	Recall from \prettyref{eq:log2_t} that $\widehat\calR_{\bt}( \hat g(T_{K-1}^{\textnormal{\scalebox{0.8}{C4.5}}})) \leq \log(2) $. Therefore,
	\begin{equation} \label{eq:v1}
	v(\|g\|_{\rm TV}+\|g\|_{\infty})  \leq 2\log(2) \frac{(\|g\|_{\rm TV}+\|g\|_{\infty})}{(\|g\|_{\rm TV}+\|g\|_{\infty})^2+8} 
	\le \frac{\log(2)\sqrt{2}}{4}
	\end{equation}
	where we used the fact that $x/(x^2+8) \leq \sqrt{2}/8$.
	Furthermore,
	\begin{equation} \label{eq:v-upper}
	v =  \frac{2(\widehat\calR_{\bt}( \hat g(T_{K-1})) - \widehat \calR_{\bt}( g)) }
	{(\|g\|_{\rm TV}+\|g\|_{\infty})^2+8}
	\leq \frac{2\log(2)}{(\|g\|_{\rm TV}+\|g\|_{\infty})^2+8} \leq \frac{\log(2)}{4}.
	\end{equation}
	Combining \prettyref{eq:v1} and \prettyref{eq:v-upper}, we have
	\begin{equation} \label{eq:taylor2-part}
	\begin{aligned}
	\frac{\exp(v(\|g\|_{\rm TV}+\|g\|_{\infty}))}{(1-v)^2} 
	&\leq
	 \frac{\exp(\log(2)\sqrt{2}/4)}{(1-\log(2)/4)^2}
	&\leq 2.
	\end{aligned}
	\end{equation}
Therefore, by substituting \prettyref{eq:v} and \prettyref{eq:taylor2-part} into \prettyref{lmm:gainv}, we have
	\begin{align*}
	\mathcal{IG}( \bt) &\geq
	v (\widehat\calR_{\bt}( \hat g(T_{K-1})) - \widehat\calR_{\bt}( g) ) - v^2 \Big(\frac{(\|g\|_{\rm TV}+\|g\|_{\infty})^2}{4}+2\Big) \\
	&= \frac{(\widehat\calR_{\bt}( \hat g(T_{K-1})) - \widehat\calR_{\bt}( g))^2}{(\|g\|_{\rm TV}+\|g\|_{\infty})^2+8} \\
	&\ge \frac{(\widehat\calR_{\bt}( \hat g(T_{K-1})) - \widehat\calR_{\bt}( g))^2}{(\|g\|_{\rm TV}+\|g\|_{\infty}+3)^2}. \qedhere
	\end{align*}
\end{proof}

\subsection{Empirical risk}

\begin{proof}[Proof of \prettyref{thm:training}]
Combining \prettyref{lmm:gain} and \prettyref{lmm:gaind} with $d=1$ for all additive $g(\cdot) \in \calG^1$ proves \prettyref{thm:training}.
\end{proof}

\subsection{Oracle inequality}

We separately prove \prettyref{thm:oracle} for CART and C4.5.

\subsubsection{Oracle inequality for CART}\label{supp:oracle_cart}
\begin{proof}[Proof of \prettyref{thm:oracle} for CART]
We first assume that $ |y_i | \leq U $, $ i = 1, 2, \dots, N $, for some $ U \geq 0 $. 
Write $ \|\hat g(T_{K}^{\textnormal{\scalebox{0.8}{CART}}}) - g^*\|^2 = E_1 + E_2 $, where
\begin{equation} \label{eq:E1}
E_1 \coloneqq \| \hat g(T_{K}^{\textnormal{\scalebox{0.8}{CART}}}) - g^*  \|^2 - 2(\|  y -\hat g(T_{K}^{\textnormal{\scalebox{0.8}{CART}}})\|_{\calD}^2 - \|  y - g^*\|_{\calD}^2) - \alpha - \beta
\end{equation}
and
$$
E_2 \coloneqq 2(\| y -\hat g(T_{K}^{\textnormal{\scalebox{0.8}{CART}}}) \|^2_{\calD}- \|  y - g^* \|_{\calD}^2) + \alpha + \beta,
$$
and $ \alpha $ and $ \beta $ are positive constants to be chosen later.
Notice that by \prettyref{thm:training}, we have
\begin{equation} \label{eq:T2}
E_2 \leq 2(\|  y - g \|_{\calD}^2 - \|  y - g^* \|_{\calD}^2) + \frac{2\|g\|^2_{\rm TV}}{K+3} + \alpha + \beta,
\end{equation}
for any $ g(\cdot) \in \calG^1$.
Taking expectations on both sides of \prettyref{eq:T2} and using $ \mathbb{E}_{\calD} (\|  y - g \|_{\calD}^2 - \|  y - g^* \|_{\calD}^2) = \|g-g^*\|^2 $ 
yields
\begin{equation} \label{eq:T2_expect}
\begin{aligned}
\mathbb{E}_{\calD}(E_2) & \leq 2\mathbb{E}_{\calD}(\|  y - g \|_{\calD}^2 - \|  y - g^* \|_{\calD}^2) +
\frac{2\|g\|^2_{\rm TV}}{K+3} + \alpha + \beta \\ & = 2\|g-g^*\|^2 + \frac{2\|g\|^2_{\rm TV}}{K+3}
+ \alpha + \beta.
\end{aligned}
\end{equation}

	To bound $ E_1 $, we first introduce a few useful concepts and definitions due to \cite{nobel1996histogram} for studying data-dependent partitions.
	Let 
	\begin{equation} \label{eq:Pi_N}
	\Pi_N\coloneqq \{ \calP(\{(\bx_1, y_1), (\bx_2, y_2), \dots, (\bx_N, y_N)\}) : (\bx_i, y_i) \in \mathbb{R}^p \times \mathbb{R} \}
	\end{equation}
	be the family of all achievable partitions $ \calP $ by growing a depth $ K $ binary tree on $ N $ points (in particular, note that $ \Pi_N$ contains all data-dependent partitions).
		Define $$ M(\Pi_N) \coloneqq \max \{ \#\calP : \calP \in \Pi_N\} $$ to be the maximum number of terminal nodes among all partitions in $ \Pi_N$. Note that $ M(\Pi_N) \leq 2^K $.
	Given a set $ \bz^N = \{\bz_1, \bz_2, \dots, \bz_N\} \subset \mathbb{R}^p $, we also define $ \Delta(\bz^N, \Pi_N) $ to be the number of distinct partitions of $ \bz^N $ induced by elements of $ \Pi_N$, that is, the number of different partitions $ \{\bz^N \cap A : A \in \calP \} $, for $ \calP \in \Pi_N$. The partitioning number $ \Delta_N(\Pi_N) $ is defined by
	$$
	\Delta_N(\Pi_N) \coloneqq \max\{ \Delta(\bz^N, \Pi_N) : \bz_1, \bz_2, \dots, \bz_N\in \mathbb{R}^p \},
	$$
	i.e., the maximum number of different partitions of any $N$ point set that can be induced by members of $ \Pi_N$.
	Finally, let $ \calG_N$ denote the collection of all piece-wise constant functions (bounded by $ U $) on partitions $ \calP \in \Pi_N$. 

Now, by \cite[Theorem 11.4]{gyorfi2002distribution} (choosing, in their notation, $ \epsilon = 1/2 $), we have
\begin{equation}
\begin{aligned} \label{eq:prob}
\mathbb{P}_{\calD}(E_1 \geq 0) &\leq \mathbb{P}_{\calD}\big(\exists \; g(\cdot) \in \calG_N: \|g-g^* \|^2 \geq 2(\|  y - g\|_{\calD}^2 - \|  y - g^*\|_{\calD}^2) + \alpha  + \beta \big)  \\ 
& \leq 14\sup_{\bx^N}\calN\Big(\frac{\beta}{40U},\calG_N, \mathscr{L}_1(\mathbb{P}_{\bx^N})\Big)\exp\Big(-\frac{\alpha N}{2568U^4}\Big),
\end{aligned}
\end{equation}
where $ \bx^N = \{\bx_1, \bx_2, \dots, \bx_N\} \subset \mathbb{R}^p $ and $ \calN(r, \calG_N, \mathscr{L}_1(\mathbb{P}_{\bx^N})) $ is the covering number for $ \calG_N$ by balls of radius $ r > 0 $ in $ \mathscr{L}_1(\mathbb{P}_{\bx^N}) $ with respect to the empirical discrete measure $ \mathbb{P}_{\bx^N} $ on $ \bx^N $.

	Next, we use \cite[Lemma 13.1 and Theorem 9.4]{gyorfi2002distribution} as in the proof of \cite[Theorem 13.1]{gyorfi2002distribution} with $ \epsilon = (32/40)\beta $ 
	to bound the empirical covering number by
	\begin{equation} \label{eq:cover}
	\calN\Big(\frac{\beta}{40U}, \calG_N, \mathscr{L}_1(\mathbb{P}_{\bx^N})\Big) \leq \Delta_N(\Pi_N)\Big(\frac{40}{32}\cdot \frac{333eU^2}{\beta}\Big)^{2M(\Pi_N)} \leq \Delta_N(\Pi_N)\Big(\frac{417eU^2}{\beta}\Big)^{2^{K+1}}.
	\end{equation}
	We further bound \prettyref{eq:cover} by noting that 
	\begin{equation} \label{eq:partition}
	\Delta_N(\Pi_N) \leq ((N-1)p)^{2^{K}-1} \leq (Np)^{2^K}.
	\end{equation} To see this,
	let $ \delta_K \coloneqq \Delta_N(\Pi_N) $ to emphasize the dependence on the depth $ K $. Now, for each depth $ K $ tree constructed from a set of $ N $ points in $ \mathbb{R}^p $, there are at most $ (N-1)p $ possible split points at the root node. The left and right (sub)trees resulting from this split have depth $ K-1 $, and hence they each independently contribute to the count in $ \delta_{K-1} $. Thus, $ \delta_K $ satisfies the recursion $ \delta_K \leq ((N-1)p)\cdot \delta_{K-1}\cdot\delta_{K-1} = ((N-1)p) \delta^2_{K-1}$ with $ \delta_1 \leq (N-1)p $. It can easily be shown that the solution is $ \delta_K \leq ((N-1)p)^{2^{K}-1} $ for $ K \geq 1 $. (The authors were inspired by \cite{scornet2015consistency} for this estimate of $ \Delta_N(\Pi_N) $.) 
	We therefore can bound the covering number by
	\begin{equation} \label{eq:covermain}
	\calN\Big(\frac{\beta}{40U}, \calG_N, \mathscr{L}_1(\mathbb{P}_{\bx^N})\Big) \leq  (Np)^{2^K}\Big(\frac{417eU^2}{\beta}\Big)^{2^{K+1}}.
	\end{equation}

Returning to \prettyref{eq:prob} and applying \prettyref{eq:covermain}
to bound the covering number, since $\hat g(T_{K}^{\textnormal{\scalebox{0.8}{CART}}})(\cdot) \in \calG_N$, we thus have
$$
\mathbb{P}_{\calD}(E_1 \geq 0) \leq 14(Np)^{2^{K}}\Big(\frac{417eU^2}{\beta}\Big)^{2^{K+1}}\exp\Big(-\frac{\alpha N}{2568U^4}\Big).
$$
We choose $\alpha \asymp \frac{U^4 2^K \log(Np)}{N}$ and $\beta \asymp \frac{U^2}{N}$
 so that $ \mathbb{P}_{\calD}(E_1 \geq 0) \leq C_1'/N $ for come universal constant $C_1'$. Furthermore, since $ E_1 \leq \| \hat g(T_{K}^{\textnormal{\scalebox{0.8}{CART}}}) -g^* \|^2 + 2\|  y - g^*\|_{\calD}^2 \leq 12U^2 $, we have $ \mathbb{E}_{\calD}(E_1) \leq 12U^2\mathbb{P}_{\calD}(E_1 \geq 0) \leq 12C_1'U^2/N $.

Adding this bound on $ \mathbb{E}_{\calD}(E_1) $ to the bound on $ \mathbb{E}_{\calD}(E_2) $ from \prettyref{eq:T2_expect} and plugging in the choices of $ \alpha $ and $ \beta $, we have
\begin{equation}
\begin{aligned} \label{eq:bounded_prediction}
\mathbb{E}_{\calD}(\|\hat g(T_{K}^{\textnormal{\scalebox{0.8}{CART}}}) - g^*\|^2) & = \mathbb{E}_{\calD}(E_1) + \mathbb{E}_{\calD}(E_2) \\ 
& \leq 2\|g-g^*\|^2 + \frac{2\|g\|^2_{\rm TV}}{K+3} + C_1''\Big(\frac{ U^42^{K}\log(Np)}{N} + \frac{U^2}{N}\Big),
\end{aligned}
\end{equation}
where $ C_1''$ is a positive universal constant.

Next, we consider the case when the response data is not bounded.
Let $ E = \bigcap_i \{ |y_i| \leq U\} $, where $ U \geq \|g^*\|_\infty $.
Using \prettyref{eq:bounded_prediction}, we obtain
\begin{equation}
\begin{aligned} \label{eq:decompose}
\mathbb{E}_{\calD}(\|\hat g(T_{K}^{\textnormal{\scalebox{0.8}{CART}}})-g^*\|^2) & =  \mathbb{E}_{\calD}(\|\hat g(T_{K}^{\textnormal{\scalebox{0.8}{CART}}}) - g^*\|^2\mathbf{1}(E)) + \mathbb{E}_{\calD}(\|\hat g(T_{K}^{\textnormal{\scalebox{0.8}{CART}}}) - g^*\|^2\mathbf{1}(E^c)) \\
& \leq 2\|g-g^*\|^2 + \frac{2\|g\|^2_{\rm TV}}{K+3} + C_1''\Big(\frac{ U^42^{K}\log(Np)}{N} + \frac{U^2}{N}\Big) \\ & \qquad + \mathbb{E}_{\calD}(\|\hat g(T_{K}^{\textnormal{\scalebox{0.8}{CART}}}) - g^*\|^2\mathbf{1}(E^c)).
\end{aligned}
\end{equation}
By a union bound, we have
\begin{equation}\label{eq:E^c}
\mathbb{P}_{\calD}(E^c) \leq N \mathbb{P}(|y| > U) \leq N \mathbb{P}(|\varepsilon| > U-\|g^*\|_\infty) \leq 2N \exp\Big(-\frac{(U-\|g^*\|_\infty)^2}{2\sigma^2}\Big),
\end{equation}
per the noise condition \prettyref{eq:bounded}.
Next, by the Cauchy-Schwarz inequality and \prettyref{eq:E^c},
\begin{equation} \label{eq:tail}
\begin{aligned}
\mathbb{E}_{\calD}(\|\hat g(T_{K}^{\textnormal{\scalebox{0.8}{CART}}}) - g^*\|^2\mathbf{1}(E^c)) &
\leq \sqrt{\mathbb{E}_{\calD}(\|\hat g(T_{K}^{\textnormal{\scalebox{0.8}{CART}}}) - g^*\|^4) \mathbb{P}_{\calD}(E^c)} \\
& \leq \sqrt{32N(\mathbb{E}_{\calD}(\|\hat g(T_{K}^{\textnormal{\scalebox{0.8}{CART}}})\|^4) +\|g^*\|^4_{\infty})} \exp\Big(-\frac{(U-\|g^*\|_\infty)^2}{4\sigma^2}\Big).
\end{aligned}
\end{equation}
Notice that $ \mathbb{E}_{\calD}(\|\hat g(T_{K}^{\textnormal{\scalebox{0.8}{CART}}})\|^4) \leq \mathbb{E}_{\calD}(\max_{i} |y_i|^4) \leq N\mathbb{E}(|y|^4) \leq 8N(\|g^*\|^4_{\infty}+\mathbb{E}(\varepsilon^4))$.
Choosing $ U = \|g^*\|_{\infty} + 2\sigma\sqrt{2\log(N)} $ yields that $ \mathbb{E}_{\calD}(\|\hat g(T_{K}^{\textnormal{\scalebox{0.8}{CART}}}) - g^*\|^2\mathbf{1}(E^c)) \leq C'''_1/N $, for some positive constant $ C'''_1 $ that depends only on $  \|g^*\|_{\infty} $ and $ \sigma^2 $. Plugging this choice of $ U$ into \prettyref{eq:tail} and then using the risk bound \prettyref{eq:decompose}, we obtain
$$
\mathbb{E}_{\calD}(\|\hat g(T_{K}^{\textnormal{\scalebox{0.8}{CART}}})-g^*\|^2) \leq 2\|g-g^*\|^2 + \frac{2\|g\|^2_{\rm TV}}{K+3} + C_1\frac{2^{K}\log^2(N)\log(Np)}{N},
$$
for some constant $ C_1 > 0 $ that depends only on $  \|g^*\|_{\infty} $ and $ \sigma^2 $.
\end{proof}

\subsubsection{Oracle inequality for C4.5} \label{supp:oracle_c4.5}

We prove \prettyref{thm:oracle} for C4.5 in two steps: we first relate the mis-classification risk to the logistic risk, and then use an empirical process argument to relate the logistic risk to the empirical risk.

Recalling \prettyref{eq:h_c},
the tree output $\hat g(T_K^{\textnormal{\scalebox{0.8}{C4.5}}})(\cdot)$ will be $ \pm \infty $ for homogenous nodes $\bt$, i.e., where $\eta_\bt = 0$ or $\eta_\bt = 1$, which means the logistic risk $\calR(\hat g(T_K^{\textnormal{\scalebox{0.8}{C4.5}}}))$ could be infinite.
To circumvent this problem, we 
define the \emph{truncated} output of a
C4.5 classification tree as
\begin{equation} \label{eq:truncated_output}
\tilde g(T_K^{\textnormal{\scalebox{0.8}{C4.5}}})(\bx) \coloneqq \log \big(\frac{\tilde\eta_{\bt}}{1-\tilde\eta_{\bt}}\big), \quad 
\tilde\eta_{\bt} \coloneqq \begin{cases}
\eta_{\bt}, &  \text{if} \; \eta_{\bt} \neq 0,1 \\
\frac{1}{N+2}, & \text{if} \; \eta_{\bt} = 0 \\
\frac{N+1}{N+2}, & \text{if} \; \eta_{\bt} = 1
\end{cases}, \quad \bx \in \bt.
\end{equation}
The following lemma relates the excess mis-classification risk to the excess logistic risk for the truncated output of the tree.

\begin{lemma} \label{lmm:pinsker}
We have
	\begin{equation} 
	\mathbb{E}_{\calD} ({\rm Err}(\hat g(T_K^{\textnormal{\scalebox{0.8}{C4.5}}})))
	- {\rm Err}(g^*) \leq \sqrt{2( \mathbb{E}_{\calD}(\calR(\tilde g(T_K^{\textnormal{\scalebox{0.8}{C4.5}}}))) - \calR(g^*)) }. 
	\end{equation}
\end{lemma}
\begin{proof}
	First notice that the truncated output of the tree makes the same class prediction as the original tree, so we have
	\begin{equation*} 
	{\rm Err}(\hat g(T_K^{\textnormal{\scalebox{0.8}{C4.5}}})) = {\rm Err}(\tilde g(T_K^{\textnormal{\scalebox{0.8}{C4.5}}})).
	\end{equation*}
	
	 Define the estimated conditional class probability of the truncated output of the tree as
	\begin{equation}\label{eq:eta_truncated}
	\tilde{\eta}(\bx) = \frac{1}{1+\exp(-\tilde g(T_K^{\textnormal{\scalebox{0.8}{C4.5}}})(\bx) )}.
	\end{equation}
	Now we first use \cite[Theorem 1.1]{gyorfi2002distribution} for plug-in classifiers, 
	to see that
	\begin{equation} \label{eq:plug_in}
	{\rm Err}(\hat g(T_K^{\textnormal{\scalebox{0.8}{C4.5}}})) -  {\rm Err}(g^*) = {\rm Err}(\tilde g(T_K^{\textnormal{\scalebox{0.8}{C4.5}}}))
	- {\rm Err}(g^*) \leq 2\mathbb{E}_{\bx}|\tilde \eta(\bx)-\eta^*(\bx)| 
	\end{equation}
	We then use Pinsker's inequality \citep[Lemma 2.5]{tsybakov2009nonparametric} 
	to show that, for each $ \bx $,
	$$
	|\tilde \eta(\bx)-\eta^*(\bx)| \leq \sqrt{\frac{1}{2}\infdiv{\eta^*(\bx)}{\tilde \eta(\bx)}},
	$$
	where $ \infdiv{p}{q} \coloneqq p\log(p/q) + (1-p)\log((1-p)/(1-q))$ is the \emph{binary relative entropy}. 
	Taking expectations and using Jensen's inequality, we get
	\begin{equation} \label{eq:pinsker}
	\mathbb{E}_{\bx}|\tilde \eta(\bx)-\eta^*(\bx)| \leq \mathbb{E}_{\bx} \sqrt{\frac{1}{2}\infdiv{\eta^*(\bx)}{\tilde \eta(\bx)}} \leq  \sqrt{\frac{1}{2}\mathbb{E}_{\bx}\infdiv{\eta^*(\bx)}{\tilde \eta(\bx)}}.
	\end{equation}
	Using $\eta^*(\bx) = \mathbb{P}(y=1\mid\bx) = 1/(1+\exp(-g^*(\bx)))$ in conjunction with \prettyref{eq:eta_truncated} we see that
	\begin{equation}\label{eq:KL}
	\begin{aligned}
	\mathbb{E}_{\bx}\infdiv{\eta^*(\bx)}{\tilde \eta(\bx)} &=
	\mathbb{E}_{\bx}\Big( \eta^*(\bx) \log \frac{1}{\tilde \eta(\bx)} + (1-\eta^*(\bx)) \log \frac{1}{1 - \tilde \eta(\bx)} \Big) \\
	&\qquad -\mathbb{E}_{\bx}\Big( \eta^*(\bx) \log \frac{1}{\eta^*(\bx)} + (1-\eta^*(\bx)) \log \frac{1}{1 - \eta^*(\bx)} \Big)\\
	&= \mathbb{E}_{(\bx,y)} \log(1+\exp(-y \tilde g(T_K^{\textnormal{\scalebox{0.8}{C4.5}}})) - \mathbb{E}_{(\bx,y)} \log(1+\exp(-y g^*))\\
	&= \calR(\tilde g(T_K^{\textnormal{\scalebox{0.8}{C4.5}}})) -\calR(g^*).
	\end{aligned}
	\end{equation}
	Combining \prettyref{eq:plug_in}, \prettyref{eq:pinsker}, and \prettyref{eq:KL} shows that
	\begin{equation} \label{eq:Err}
	{\rm Err}(\hat g(T_K^{\textnormal{\scalebox{0.8}{C4.5}}}))
	- {\rm Err}(g^*) \leq \sqrt{2 (\calR(\tilde g(T_K^{\textnormal{\scalebox{0.8}{C4.5}}})) -\calR(g^*) )}. 
	\end{equation}
	Taking expectations of both sides of \prettyref{eq:Err}
	and using Jensen's inequality 
	 on the right hand side completes the proof.	
\end{proof}

For the second step, we use an empirical process argument to argue that the empirical risk of the truncated output of the tree concentrates near the logistic risk of the truncated output of the tree.

\begin{lemma}\label{lmm:concentration}
	For any $ \delta > 0 $,
	we have
	$$
	\mathbb{P}_{\calD} (| \widehat \calR( \tilde g(T_K^{\textnormal{\scalebox{0.8}{C4.5}}})) - \calR(\tilde g(T_K^{\textnormal{\scalebox{0.8}{C4.5}}})) | > \delta) \geq  1- 8(N^3p)^{2^K}\exp(-N \delta^2/(128\log^2(N+2))).
	$$
\end{lemma}

\begin{proof}
We use concepts that were introduced in the proof of \prettyref{thm:oracle} for CART in \prettyref{supp:oracle_cart}.
Let $ \widetilde \calG_N$ denote the collection of all piece-wise constant functions on a partition $ \calP \in \Pi_N$ ($\Pi_N$ is defined in  \prettyref{eq:Pi_N}), where each of the constants has the form $ \log(a/(b-a)) $, with $ b > a $ and $ a,b \in \{1, 2, \dots, N\} $.
Let $ \calH $ denote the set of all mappings $ h(\bz) \coloneqq \log(1+\exp(-y \tilde g(\bx))$, where $ \bz = (\bx, y) \in \mathbb{R}^p \times \{-1, 1\} $ and $ \tilde g(\cdot) \in \calG_N$.
We have
$$
\mathbb{P}_{\calD} (| \widehat \calR( \tilde g(T_K^{\textnormal{\scalebox{0.8}{C4.5}}})) - \calR(\tilde g(T_K^{\textnormal{\scalebox{0.8}{C4.5}}})) | > \delta)  \leq \mathbb{P}_{\calD}\Big( \sup_{h(\cdot) \in \calH}\Big|\frac{1}{N}\sum_{i=1}^N h(\bz_i) - \mathbb{E}_{\bz}h(\bz) \Big| > \delta \Big).
$$
	Since $ h(\bz) \in [0, \log(N+2)] $ for all $ \bz $, by \cite[Theorem 9.1]{gyorfi2002distribution}, we have that
	\begin{align*}
	&\mathbb{P}_{\calD}\Big( \sup_{h(\cdot) \in \calH}\Big|\frac{1}{N}\sum_{i=1}^N h(\bz_i) - \mathbb{E}_{\bz}h(\bz) \Big| > \delta \Big) \\ 
	& \qquad \leq 8 \sup_{\bz^N} \calN(\delta/8, \calH, \mathscr{L}_1(\mathbb{P}_{\bz^N}))  \exp\Big(-\frac{N\delta^2}{128\log^2(N+2)}\Big).
	\end{align*}
Next, we show that, for all $ \delta \geq  0$,
    \begin{equation} \label{eq:covering}
	\calN(\delta, \calH, \mathscr{L}_1(\mathbb{P}_{\bz^N})) \leq (N^3p)^{2^K}.
     \end{equation}
   
Fix $ \bz^N = \{(\bx_1, y_1), (\bx_2, y_2), \dots, (\bx_N, y_N)\} = \{ \bz_1,  \bz_2, \dots, \bz_N \} \subset  \mathbb{R}^p \times \{-1, 1\} $.  
We aim to bound the cardinality 
of the set
\begin{equation} \label{eq:set_output}
\{ h(\bz) : \bz \in \bz^N, \; h(\cdot) \in \calH\} = \{ \log(1+\exp(-y \tilde g(\bx_i))) :  (\bx_i, y_i) \in \bz^N, \; \tilde g(\cdot) \in \widetilde \calG_N \}.
\end{equation}
To this end, note that a given $\calP \in \Pi_N$ induces a partition 
of $\bx^N = \{ \bx_1, \dots, \bx_N \}  $ consisting of sets $B = \bx^N \cap A $, where $ A $ is an element of $ \calP $.
For all $ \bx \in B $, there are at most $ N^2 $ possible outputs $\log(a/(b-a))$ for a piece-wise constant function $\tilde g(\cdot) \in \widetilde \calG_N$ on $\calP$.
Since there are at most $M(\Pi_N) \leq 2^K $ sets $B$, the number of possible outputs at the points $\bx^N$ 
is at most $(N^2)^{2^K}$.
Any other $\calP'\in \Pi_N$ that induces the same partition of $ \bx^N $ as $ \calP $ will result in the same set of possible outputs for $\tilde g(\cdot)$ at $\bx^N$.
Since are at most $\Delta_N(\Pi_N) \leq (p(N-1))^{2^K-1}$ induced partitions $\{\bx^N \cap A: A \in \calP\}$ (see \prettyref{eq:partition}), the cardinality of \prettyref{eq:set_output}
is at most $  (p(N-1))^{2^K-1} (N^2)^{2^K} \leq (N^3p)^{2^K} $.
Since the cardinality of the set \prettyref{eq:set_output} upper bounds $ \calN(\delta, \calH, \mathscr{L}_1(\mathbb{P}_{\bz^N})) $ for all $ \delta \geq 0 $, we have established \prettyref{eq:covering} and hence \prettyref{lmm:concentration} as well.
\end{proof}

Now we are ready to complete the proof of \prettyref{thm:oracle}.

\begin{proof}[Proof of \prettyref{thm:oracle}]
In light of \prettyref{lmm:pinsker}, it suffices to upper bound 
\begin{equation} \label{eq:decomposition}
\begin{aligned}
\mathbb{E}_{\calD}(\calR(\tilde g(T_K^{\textnormal{\scalebox{0.8}{C4.5}}}))) - \calR(g^*) &\leq
\underbrace{\mathbb{E}_{\calD}(\calR(\tilde g(T_K^{\textnormal{\scalebox{0.8}{C4.5}}})) - \widehat\calR(\tilde g(T_K^{\textnormal{\scalebox{0.8}{C4.5}}})))}_{\text{(I)}}
+\underbrace{\mathbb{E}_{\calD}(\widehat \calR(\tilde g(T_K^{\textnormal{\scalebox{0.8}{C4.5}}})) - \widehat\calR(\hat g(T_K^{\textnormal{\scalebox{0.8}{C4.5}}})))}_{\text{(II)}} \\
& \quad + \underbrace{\mathbb{E}_{\calD}(\widehat\calR(\hat g(T_K^{\textnormal{\scalebox{0.8}{C4.5}}})) )- \calR(g)}_{\text{(III)}} + \underbrace{\calR(g) - \calR(g^*)}_{\text{(IV)}}.
\end{aligned}
\end{equation}
We now separately bound the four terms in \prettyref{eq:decomposition}.
By \prettyref{lmm:concentration},
for any $\delta >0$ we have
$$
\text{(I)} 
\leq \delta + 8(N^3p)^{2^K}\exp\Big(-\frac{N \delta^2}{128\log^2(N+2)}\Big) \log(N+2),
$$
so setting
$ \delta^2 \asymp \frac{2^K \log^2(N)\log(Np)}{N} $ 
shows that
\begin{equation}\label{eq:I}
\text{(I)} \leq C_2' \sqrt{\frac{2^K\log^2(N)\log(Np)}{N}},
\end{equation}
for some positive universal constant $C_2'$.
For the second term, notice that \prettyref{eq:truncated_output} says that $\tilde g(T_K^{\textnormal{\scalebox{0.8}{C4.5}}})$ only differs from $\hat g(T_K^{\textnormal{\scalebox{0.8}{C4.5}}})$ when $\eta_\bt = 0$ or $\eta_{\bt} = 1$, so we have
\begin{equation} \label{eq:II}
\text{(II)} \leq \log(1+\exp (-\log(N+1))) 
\leq \frac{1}{N+1}.
\end{equation}
 \prettyref{thm:training} allows us to bound the third term by
\begin{equation} \label{eq:III}
\text{(III)} = \mathbb{E}_{\calD}(\widehat\calR(\hat g(T_K^{\textnormal{\scalebox{0.8}{C4.5}}}))-\widehat\calR(g)) \leq \frac{(\|g\|_{\rm TV} + \|g\|_\infty+3)^2}{K+3}.
\end{equation}
Finally for the last term, we use a second order Taylor expansion of the logistic function \prettyref{eq:taylor} to see that
\begin{equation*}
\begin{aligned}
\text{(IV)} 
&= \mathbb{E}_{(\bx,y)}(q(y g(\bx)) - q(yg^*(\bx))) \\
&\leq \mathbb{E}_{(\bx,y)}\Big(y( g(\bx) - g^*(\bx))q'(yg^*(\bx)) + \frac{(g(\bx)-g^*(\bx))^2}{8}\Big),
\end{aligned}
\end{equation*}
where we used a uniform bound on the second derivative \prettyref{eq:q' and q''}. But notice that
\begin{align*}
&\mathbb{E}_{(\bx,y)} (y( g(\bx) - g^*(\bx))q'(yg^*(\bx)))\\
& \qquad=  \mathbb{E}_{\bx} \Big(- \eta^*(\bx)(g(\bx) - g^*(\bx)) \frac{e^{-g^*(\bx)}}{1+e^{-g^*(\bx)}}  + (1-\eta^*(\bx)) (g(\bx) - g^*(\bx)) \frac{e^{g^*(\bx)}}{1+e^{g^*(\bx)}}\Big) \\
& \qquad =0,
\end{align*}
where in the last line we used $\eta^*(\bx) = 1/(1+\exp(-g^*(\bx)))$.
Thus we have shown that
\begin{equation}\label{eq:IV}
\text{(IV)} \leq \frac{\|g- g^*\|^2}{8}.
\end{equation}
Combining \prettyref{eq:decomposition}, \prettyref{eq:I}, \prettyref{eq:II}, \prettyref{eq:III}, \prettyref{eq:IV} and \prettyref{lmm:pinsker} shows that
\begin{equation} \label{eq:oracle2}
\begin{aligned}
&\mathbb{E}_{\calD} ({\rm Err}(\hat g(T_K^{\textnormal{\scalebox{0.8}{C4.5}}})))
- {\rm Err}(g^*) \\
& \qquad\qquad \leq \sqrt{\frac{\|g - g^*\|^2}{4} + \frac{2(\|g\|_{{\rm TV}}+\|g\|_\infty+3)^2}{K+3} 
	+\frac{2}{N+1}
	+ 2C_2'\sqrt{\frac{2^K\log^2(N)\log(Np)}N}}\\
&\qquad\qquad \leq \|g-g^*\|+ 2\frac{\|g\|_{{\rm TV}}+\|g\|_\infty+3}{\sqrt{K+3}} +C_2\Big(\frac{2^K\log^2(N)\log(Np)}{N}\Big)^{1/4},
\end{aligned}
\end{equation}
where $C_2$ is a positive universal constant. 
\end{proof}

\subsection{Consistency for unbounded variation component functions} \label{supp:consistent_general}
\begin{proof}[Proof of \prettyref{cor:consistent_general}]
By \citep[Theorem 4.3]{stein2005real}, there exists a sequence of integrable step functions $ (s_{jN}(\cdot))_N$ that converges pointwise almost surely to $ g_j(\cdot) $. (The proof of \citep[Theorem 4.3]{stein2005real} extends to any Borel probability measure on $ \mathbb{R}^p $.) By expanding the sequence to include successive duplicate elements, we can assume without loss of generality that $ \|s_{jN}\|_{\rm TV} = o(\sqrt{K_N}) $. 
Next, define $ s_N(\bx) \coloneqq \sum_{j=1}^{p_N} s_{jN}(x_j) $ so that by the Lebesgue dominated convergence theorem and the fact that $ \sup_N\|g_N^*\|_{\ell_0} < \infty $, we have $ \lim_{N\rightarrow\infty}\|s_N-g^*_N\| = 0 $. Furthermore, since each component $ s_{jN}(\cdot) $ is a step function with total variation $ o(\sqrt{K_N}) $, we likewise have $ \|s_N\|_{\rm TV} = o(\sqrt{K_N}) $, where we again used the assumption that $ \sup_N\|g_N^*\|_{\ell_0} < \infty $. Thus, by \prettyref{thm:oracle} and the hypotheses of \prettyref{cor:consistent}, it follows that
\begin{align*}
\mathbb{E}_{\calD}(\| \hat g(T_{K_N}^{\textnormal{\scalebox{0.8}{CART}}}) - g^*_N\|^2) \leq
2\|s_N - g_N^*\|^2 + \frac{2\|s_N\|^2_{\rm TV}}{K_N+3} + 2C_1\frac{2^{K_N}\log^2(N) \log(Np_N)}{N} \rightarrow 0,
\end{align*}
as $N \rightarrow \infty$. Similarly, it can be shown that
\begin{equation*}
\lim_{N \rightarrow\infty}(\mathbb{E}_{\calD}({\rm Err}(\hat g(T_{K_N}^{\textnormal{\scalebox{0.8}{C4.5}}}))) - {\rm Err}(g^*_N)) \to 0, \quad \text{as} \quad N \rightarrow \infty. \qedhere
\end{equation*}
\end{proof}

\section{Beyond discriminative models}
We prove the results of \prettyref{sec:margin}, namely, \prettyref{lmm:gain_margin} for the impurity gain, \prettyref{thm:training_margin} for the empirical risk, and \prettyref{thm:oracle_margin} for the mis-classification risk in the large margin setting.

\subsection{Information gain for C4.5 with large margin} \label{supp:gain_margin}

We prove \prettyref{lmm:gain_margin} following a similar strategy as the proof of \prettyref{lmm:gain} for C4.5 in \prettyref{supp:gain_id3}.

We consider the same empirical measure as \prettyref{eq:prior_id3}
but replace $g(\cdot)$ with $f^*(\cdot)$:
$$
\frac{d\Pi(j, s)}{d(j,s)} \coloneqq \frac{|D f_j(s)|}{\sum_{j'=1}^p \int |Df_{j'}(s')|ds'}.
$$
Analogous to \prettyref{eq:prior} and \prettyref{eq:TV}, we have
\begin{equation} \label{eq:prior2}
\int \text{sgn}(Df_j(s))\mathbf{1}(x_{ij} > s) d\Pi(j, s)
=\frac{f^*(\bx_{i})-  f^*(\bx_{(1)}) }{\sum_{j'=1}^p \int |Df_{j'}(s')|ds'}
\end{equation}
and
\begin{equation} \label{eq:TV2}
\sum_{j'=1}^p \int |Df_{j'}(s')|ds' 
\leq \|f^*\|_{\rm TV} \leq 1,
\end{equation}
where we used \prettyref{ass:margin}.
Before proving \prettyref{lmm:gain_margin}, we prove the following statement analagous to \prettyref{lmm:gainv}.
\begin{lemma} \label{lmm:gainv_margin}
For any $v \in [0,1]$ and any terminal node $\bt$ of $T_{K-1}^{\textnormal{\scalebox{0.8}{C4.5}}}$, we have 
\begin{align*}
\mathcal{IG}(\bt) \ge 2v(\gamma- ve^{2v})\eta_{\bt}(1-\eta_{\bt}).
\end{align*}
\end{lemma}
\begin{proof}
	We slightly modify the definition of the decision stump in \prettyref{eq:beta_stump}.
	Let
	\begin{align*}
	&\beta_{\bt}(i,j, s) = \beta_{\bt} + \beta(f^*) \mathbf{1}(x_{ij}>s), \\
	&\beta(f^*) = v  
	\Big(
	\text{sgn}(Df_j(s))\sum_{j'=1}^p \int |Df_{j'}(s')|ds' 
	+ f^*(\bx_{(1)}) \Big).
	\end{align*}
	By \prettyref{eq:TV2} and \prettyref{ass:margin}, we have
	\begin{equation} \label{eq:beta}
	|\beta(f^*)| \le v \Big(\sum_{j'=1}^p \int |Df_{j'}(s')|ds'+ |f^*(\bx_{(1)})| \Big) \leq v(1+\|f^*\|_\infty) \le 2v.
	\end{equation}
	
Similar to \prettyref{eq:IG-lower} and \prettyref{eq:taylor2}, we have
	\begin{equation} \label{eq:IG-lower-better}
	\begin{aligned}
	\mathcal{IG}( \bt) 
	&\ge \frac{1}{N_{\bt}}\sum_{\bx_i \in \bt}  \int  (q(y_i \beta_{\bt}) - q(y_i\beta_{\bt}(i,j, s)))d\Pi(j, s) \\
	& \ge - \frac{1}{N_{\bt}}\sum_{\bx_i \in \bt}  \Big(\int  y_i\beta(f^*)  \mathbf{1}(x_{ij}>s) q'(y_i \beta_{\bt})d\Pi(j, s) + 2v^2 q''(c_i) \Big),\\
	\end{aligned}
	\end{equation}
	where $c_i$ is between $y_i\beta_{\bt}$ and $y_i\beta_{\bt}(i,j, s)$
	 and we used the fact that $|y_i \beta_\bt - y_i \beta_\bt(i,j,s)| = |\beta(f^*)| \le 2v$ from \prettyref{eq:beta}. 
	 Using our expression \prettyref{eq:beta_t} for $\beta_\bt$, we can simplify both the first and second derivatives. 
	 For the second derivative, we have the uniform bound
	 \begin{equation} \label{eq:q''}
	 q''(c_i) = \frac{e^{-c_i}}{(1+e^{-c_i})^2}
	 \le e^{2v}\frac{e^{-y_i\beta_{\bt}}}{(1+e^{-y_i\beta_{\bt}})^2}  
	  = e^{2v} \eta_{\bt}(1-\eta_{\bt}).
	 \end{equation}
	 
	The first derivative also simplifies nicely:
	$$
	-q'(y_i\beta_{\bt}) = \frac{e^{-y_i\beta_{\bt}}}{1+e^{-y_i\beta_{\bt}}} = 
	\begin{cases}
	1-\eta_{\bt} & \text{if }y_i = 1 \\
	\eta_{\bt} & \text{if }y_i = -1.\\
	\end{cases}
	$$
	Thus, we split the summation into two cases depending on whether $y_i = 1$ or $y_i = -1$: 
	\begin{equation} \label{eq:y_i=1}
	\begin{aligned}
	- \frac{1}{N_{\bt}}\sum_{x_i \in \bt \atop y_i = 1}  \int y_i \beta(f^*) \mathbf{1}(x_{ij}>s) q'(y_i\beta_{\bt}) d\Pi(j,s)
	&=\frac{1-\eta_{\bt} }{N_{\bt}}\sum_{\bx_i \in \bt \atop y_i = 1}  vy_i f^*(\bx_i)  \\
	&\ge\frac{1-\eta_{\bt}}{N_{\bt}}\sum_{\bx_i \in \bt \atop y_i = 1}  v \gamma \\
	&=v \gamma (1-\eta_{\bt})\eta_{\bt},
	\end{aligned}
	\end{equation}
	and similarly
	\begin{equation} \label{eq:y_i=-1}
	\begin{aligned}
	- \frac{1}{N_{\bt}}\sum_{x_i \in \bt \atop y_i = -1}  \int y_i \beta(f^*)\mathbf{1}(x_{ij}>s) q'(y_i\beta_{\bt}) d\Pi(j,s)
	&=\frac{\eta_{\bt} }{N_{\bt}}\sum_{\bx_i \in \bt \atop y_i = -1}  vy_i f^*(\bx_i)  \\
	&\ge\frac{\eta_{\bt}}{N_{\bt}}\sum_{\bx_i \in \bt \atop y_i = -1}  v \gamma \\
	&=v \gamma (1-\eta_{\bt})\eta_{\bt}.
	\end{aligned}
	\end{equation}
	Substituting \prettyref{eq:q''}, \prettyref{eq:y_i=1}, and \prettyref{eq:y_i=-1} into \prettyref{eq:IG-lower-better} completes the proof.
	\end{proof}
	
	Now we are ready to complete the proof of \prettyref{lmm:gain_margin}.
	
	\begin{proof} [Proof of \prettyref{lmm:gain_margin}]

	Choose $ v = \gamma/4$ in \prettyref{lmm:gainv_margin} to see that 
	\begin{equation}\label{eq:IG_eta}
	\mathcal{IG}(\bt) \ge
	\Big(\frac{\gamma^2}{2} - \frac{\gamma^2\exp({\frac{\gamma}{2}}) }{8} \Big)\eta_{\bt}(1-\eta_{\bt}) \ge \frac{\gamma^2(4-\sqrt{e})}{8}\eta_{\bt}(1-\eta_{\bt}),
	\end{equation}
	since $\gamma \le 1$.
	Recalling \prettyref{eq:log2_t},
	we have that
	\begin{align*}
	\frac{\eta_{\bt}(1-\eta_{\bt})\log(1/\widehat\calR_\bt(\hat g(T_{K-1}^{\textnormal{\scalebox{0.8}{C4.5}}})))}{\widehat\calR_\bt(\hat g(T_{K-1}^{\textnormal{\scalebox{0.8}{C4.5}}}))} 
	&= 	\frac{\eta_{\bt}(1-\eta_{\bt})\log(\eta_{\bt} \log(\frac{1}{\eta_{\bt}}) + (1 - \eta_{\bt}) \log(\frac{1}{1-\eta_\bt}))}{\eta_{\bt} \log(\eta_{\bt}) +(1 - \eta_{\bt}) \log(1-\eta_\bt)},
	\end{align*}
	which is a convex function
	with minimum at $\eta_{\bt} = 1/2$; hence
	\begin{equation} \label{eq:eta_R}
	\frac{\eta_{\bt}(1-\eta_{\bt})\log(1/\widehat\calR_\bt(\hat g(T_{K-1}^{\textnormal{\scalebox{0.8}{C4.5}}})))}{\widehat\calR_\bt(\hat g(T_{K-1}^{\textnormal{\scalebox{0.8}{C4.5}}}))} 
	\ge \frac{\frac{1}{2} \cdot \frac{1}{2} \cdot \log(\log(2))}{\log(1/2)}
	\ge \frac{1}{8}.
	\end{equation}
	Combining \prettyref{eq:IG_eta} and \prettyref{eq:eta_R} 
	yields 
	\begin{equation*}
	\mathcal{IG}(\bt) \geq \frac{\gamma^2(4-\sqrt{e})}{64} \cdot \frac{\widehat\calR_\bt(\hat g(T_{K-1}^{\textnormal{\scalebox{0.8}{C4.5}}}))}{\log(1/\widehat\calR_\bt(\hat g(T_{K-1}^{\textnormal{\scalebox{0.8}{C4.5}}})))} \ge \frac{\gamma^2}{30} \cdot \frac{\widehat\calR_\bt(\hat g(T_{K-1}^{\textnormal{\scalebox{0.8}{C4.5}}}))}{\log(1/\widehat\calR_\bt(\hat g(T_{K-1}^{\textnormal{\scalebox{0.8}{C4.5}}})))}. \qedhere
	\end{equation*}
\end{proof}

\subsection{Empirical risk  for C4.5 with large margin}

As with the proof of \prettyref{lmm:gaind}, we prove \prettyref{thm:training_margin} using \prettyref{lmm:gain_margin}.

\begin{proof}[Proof of \prettyref{thm:training_margin}]
Let $A \coloneqq \gamma^2/30 $ be the constant in \prettyref{lmm:gain_margin}.
Plugging \prettyref{lmm:gain_margin}
into \prettyref{eq:training_recursion} we see that
\begin{equation} \label{eq:L-recursion}
\begin{aligned}
\widehat \calR(\hat g(T_{K}^{\textnormal{\scalebox{0.8}{C4.5}}})) &\le \widehat\calR(\hat g(T_{K-1}^{\textnormal{\scalebox{0.8}{C4.5}}})) - A\sum_{\bt \in T_{K-1}^{\textnormal{\scalebox{0.8}{C4.5}}}} \frac{N_\bt}{N}  \frac{\widehat \calR_\bt(\hat g(T_{K-1}^{\textnormal{\scalebox{0.8}{C4.5}}}))}{\log(1/ \widehat\calR_\bt(\hat g(T_{K-1}^{\textnormal{\scalebox{0.8}{C4.5}}})))} \\ 
&\le \widehat\calR(\hat g(T_{K-1}^{\textnormal{\scalebox{0.8}{C4.5}}})) - A  \frac{\sum_{\bt \in T_{K-1}^{\textnormal{\scalebox{0.8}{C4.5}}}} \frac{N_\bt}{N}\widehat\calR_\bt(\hat g(T_{K-1}^{\textnormal{\scalebox{0.8}{C4.5}}}))}{\log(1/\sum_{\bt \in T_{K-1}^{\textnormal{\scalebox{0.8}{C4.5}}}} \frac{N_\bt}{N} \widehat\calR_\bt(\hat g(T_{K-1}^{\textnormal{\scalebox{0.8}{C4.5}}})))} \\ 
&= \widehat\calR(\hat g(T_{K-1}^{\textnormal{\scalebox{0.8}{C4.5}}})) \Big(1 - \frac{A}{\log(1/\widehat\calR(\hat g(T_{K-1}^{\textnormal{\scalebox{0.8}{C4.5}}})))}\Big),
\end{aligned}
\end{equation}
where
the second inequality comes from Jensen's inequality and the fact that $ x \mapsto x/\log(x) $ is convex for $x \in [0, 1) $. 
Letting $\calE_K = \widehat\calR(\hat g(T_{K}^{\textnormal{\scalebox{0.8}{C4.5}}}))$, we end up with the recursion
\begin{equation}
\label{eq:R-recursion}
\calE_K \leq \calE_{K-1}\Big(1 - \frac{A}{\log(1/\calE_{K-1})} \Big), \quad K \geq 1.
\end{equation}

We prove that the recursion \prettyref{eq:R-recursion} has solution
$$
\widehat\calR(\hat g(T_{K}^{\textnormal{\scalebox{0.8}{C4.5}}})) = \calE_K \leq \exp(- (AK)^{1/2}).
$$
The base case follows since $R_0\le \log(2) \le 1$.
Next assume that for some $K \ge 1$ that
\begin{equation} \label{eq:induction-ass}
\calE_{K-1} \leq \exp(- (A(K-1))^{1/2}).
\end{equation}
First notice that if $\calE_{K-1} \leq \exp(- (AK)^{1/2})$, then by \prettyref{eq:R-recursion} we have that
$
\calE_K \leq \calE_{K-1} \leq \exp(- (AK)^{1/2}),
$
so our induction step is complete. Thus we may assume that
\begin{equation} \label{eq:induction-alt}
\calE_{K-1} \geq \exp(- (AK)^{1/2}).
\end{equation}
By \prettyref{eq:induction-alt} and \prettyref{eq:induction-ass}, we have that
\begin{equation} \label{eq:above}
\begin{aligned}
\calE_K &\le \calE_{K-1} \Big(1 - \frac{A }{\log(1/\calE_{K-1})} \Big) \\
&\leq \exp(- (A(K-1))^{1/2}) \Big(1 - \frac{A }{ (A K)^{1/2}} \Big) \\
&\leq \exp\Big(- (A(K-1))^{1/2} - \Big(\frac{A}{K}\Big)^{1/2} \Big).
\end{aligned}
\end{equation}
We have $(K-1)^{1/2} \ge 
K^{1/2} 
- \frac{1}{K^{1/2}} $ for $K\ge 1$,
so
\prettyref{eq:above}
is upper bounded by
$$
\exp\Big(- A^{1/2} \Big(K^{1/2} - \frac{1}{K^{1/2}} + \frac{1}{ K^{1/2}} \Big) \Big) =  \exp(-(AK)^{1/2}),
$$
which completes the induction step.
\end{proof}

\subsection{Mis-classification risk of C4.5 with large margin} \label{supp:consistent_margin}

\begin{proof}[Proof of \prettyref{thm:oracle_margin}]
The proof is similar to the proof of \prettyref{thm:oracle} for C4.5 in \prettyref{supp:oracle_c4.5}, except that we invoke a different bound on the empirical risk.
Recall that the Bayes risk ${\rm Err}(g^*)$ and population risk $\calR(g^*)$ are both zero, due to 
\prettyref{ass:margin}.
Therefore, \prettyref{lmm:pinsker} now becomes
$$
	\mathbb{E}_{\calD} ({\rm Err}(\hat g(T_K^{\textnormal{\scalebox{0.8}{C4.5}}}))) \leq \sqrt{2\mathbb{E}_{\calD}(\calR(\tilde g(T_K^{\textnormal{\scalebox{0.8}{C4.5}}}))) }.
$$
We modify \prettyref{eq:decomposition} to
\begin{equation*} 
\begin{aligned}
\mathbb{E}_{\calD}(\calR(\tilde g(T_K^{\textnormal{\scalebox{0.8}{C4.5}}}))) &\leq
\underbrace{\mathbb{E}_{\calD}(\calR(\tilde g(T_K^{\textnormal{\scalebox{0.8}{C4.5}}})) - \widehat\calR(\tilde g(T_K^{\textnormal{\scalebox{0.8}{C4.5}}})))}_{\text{(I)}}
+\underbrace{\mathbb{E}_{\calD}(\widehat \calR(\tilde g(T_K^{\textnormal{\scalebox{0.8}{C4.5}}})) - \widehat\calR(\hat g(T_K^{\textnormal{\scalebox{0.8}{C4.5}}})))}_{\text{(II)}} \\
& \quad + \underbrace{\mathbb{E}_{\calD}(\widehat\calR(\hat g(T_K^{\textnormal{\scalebox{0.8}{C4.5}}})) )}_{\text{(III)}}.
\end{aligned}
\end{equation*}
We still have the same upper bounds \prettyref{eq:I} and \prettyref{eq:II} for $\text{(I)}$ and $\text{(II)}$, respectively, since \prettyref{lmm:pinsker} and \prettyref{lmm:concentration} still hold. The key difference is that we replace \prettyref{thm:training} with \prettyref{thm:training_margin} so that
$$
\text{(III)} \leq \exp\Big(- \Big(\frac{\gamma^2K}{30} \Big)^{1/2}\Big).
$$

Analogous to \prettyref{eq:oracle2}, we have
\begin{align*}
\mathbb{E}_{\calD} ({\rm Err}(\hat g(T_K^{\textnormal{\scalebox{0.8}{C4.5}}}))) 
&\leq \sqrt{2 \exp\Big(- \Big(\frac{\gamma^2K}{30} \Big)^{1/2}\Big)+
\frac{2}{N+1} +2C_2'\sqrt{\frac{2^K\log^2(N)\log(Np)}{N}}}\\
&\leq 2 \exp\Big(- \Big(\frac{\gamma^2K}{120} \Big)^{1/2}\Big)+
	C_2\Big(\frac{2^K\log^2(N)\log(Np)}{N}\Big)^{1/4},
\end{align*}
where $C_2$ is the constant in the statement of \prettyref{thm:oracle}. 
\end{proof}

\section{Models with interactions} \label{supp:non-additive}

We prove a generalization of the empirical risk bound in \prettyref{thm:training} to the setting in \prettyref{sec:beyond_additive} where $\calG$ may have interaction terms.

\begin{lemma} \label{lmm:gaind}
	
	Let $g(\cdot) \in \calG^d$ and $ K \geq d $. Assume that for any terminal node $ \bt $ of the tree $T_{K-d} $ such that $ \widehat \calR_\bt(\hat g(T_{K-d})) > \widehat \calR_\bt (g) $, we have
	\begin{equation} \label{eq:gain_d}
	\mathcal{IG}_d(\bt)\geq \frac{(\widehat \calR_\bt(\hat g(T_{K-d})) - \widehat \calR_\bt (g))^2}{V^2(g)},
	\end{equation}
	for some complexity constant $ V(g)$ that depends only on $ g(\cdot) $. Then 
	$$
	\widehat \calR(\hat g(T_K))\leq 
	\widehat \calR(g) + 
	\frac{V(g)d}{K+2d+1}.
	$$
\end{lemma}

\begin{proof}[Proof of \prettyref{lmm:gaind}]
	Recall the notation 
	\begin{equation} 
		\calE_K \coloneqq \widehat \calR(\hat g(T_K)) - \widehat \calR(g), \quad \calE_K(\bt) \coloneqq \widehat \calR_\bt(\hat g(T_K)) - \widehat \calR_\bt(g)
	\end{equation}
	defined in \prettyref{eq:calE} for the global and within-node excess empirical risks, respectively.		
	It can be seen from \prettyref{eq:training_recursion} that the empirical risk decreases with the depth so that $  \calE_1 \geq  \calE_2  \geq \cdots \geq  \calE_K$. If $ \calE_{K-d}< 0 $, then there is nothing to prove because $  \calE_K\leq \calE_{K-d}< 0 $ and hence $\widehat \calR(\hat g(T_K)) < \widehat\calR(g)$, as desired.
	Therefore, we assume throughout that $ \calE_{K-d}\geq 0 $. 
	
	Substituting \prettyref{eq:gain_d} into \prettyref{eq:training_recursion2} and subtracting $\widehat{\calR}(g)$ from both sides, we see that
	\begin{equation} \label{eq:training_recursion_E_d}
	\calE_K \leq \calE_{K-d}- \frac{1}{V(g)}\sum_{\bt\in T_{K-d}:\calE_{K-d}(\bt) > 0}\frac{N_\bt}N \calE^2_{K-d}(\bt).
	\end{equation}
	We note that if a node $ \bt \in T_{K-d} $ cannot be refined any further, corresponding to a case where the node contains a single data point or where all input values and/or all response values within the node are the same, then $ \calE_{K-d}(\bt) \leq 0 $.
	
	By convexity of the square function and the fact that $ \sum_{\bt\in T_{K-d}: \calE_{K-d}(\bt)> 0}\frac{N_{\bt}}{N} \leq 1 $, we have from Jensen's inequality that
	\begin{equation}\label{eq:lower1}
	\begin{aligned}
	\sum_{\bt\in T_{K-d}: \calE_{K-d}(\bt)> 0}\frac{N_{\bt}}{N} \calE^2_{K-d}(\bt) &\geq \Big(\sum_{\bt\in T_{K-d}: \calE_{K-d}(\bt)> 0}\frac{N_{\bt}}{N} \calE_{K-d}(\bt)\Big)^2 \\
	&\ge 
	\Big(\sum_{\bt\in T_{K-d}}\frac{N_{\bt}}{N} \calE_{K-d}(\bt)\Big)^2\\
	&= \calE^2_{K-d},
	\end{aligned}
	\end{equation}
	where the last inequality comes from the fact that $\calE_{K-d}$ is non-negative by assumption. 
	Applying \prettyref{eq:lower1} to \prettyref{eq:training_recursion_E_d}, we obtain
	\begin{equation} \label{eq:recursion_plus}
	\calE_K\leq \calE_{K-d} \Big(1- \frac{\calE_{K-d}}{V(g)}\Big), \quad K \geq d.
	\end{equation}
	Now we prove that  $\calE_K\leq \frac{V(g)d}{K+2d+1}$ for all $K \geq d $ by induction.
	The base case $K <2d$ is established by noticing that 
	$$
	\calE_K \leq \calE_{K-d}\Big(1-\frac{\calE_{K-d}}{V(g)}\Big) \leq \frac{V(g)}{4} \leq \frac{V(g)d}{K+2d+1}.
	$$

	For $ K \ge 2d $, assume that $ \calE_{K-d}\leq \frac{V(g)d}{K+d+1}$. Then either $ \calE_{K-d}\leq \frac{V(g)d}{K+2d+1}$, in which case we are done since $  \calE_K\leq \calE_{K-d}$, or $ \calE_{K-d}> \frac{V(g)d}{K+2d+1}$, in which case
	\begin{equation*}
	\calE_K\leq \calE_{K-d}\Big(1-\frac{ \calE_{K-d}}{V(g)}\Big) \leq  \frac{V(g)d}{K+d+1} \Big(1 - \frac{d}{K+2d+1} \Big) = \frac{V(g)d}{K+2d+1}. \qedhere
	\end{equation*}
\end{proof}

\section{Random forests}

We prove the oracle inequality for random forests (\prettyref{thm:oracle_forest}) given in \prettyref{sec:forests}.

\begin{proof}[Proof of \prettyref{thm:oracle_forest}]
	The proof follows similar lines as \prettyref{thm:training} and \prettyref{thm:oracle}, but with some interesting twists due to the additional randomness in the trees.

	Throughout the proof, we focus on a single random base regression tree $T_K^{\textnormal{\scalebox{0.8}{CART}}} =T_K^{\textnormal{\scalebox{0.8}{CART}}}(\Theta) $ in the forest.
We let $\calI' \subset \{1,\ldots,N\}$ denote the random indices corresponding to the subsampled data $\calD'$ from the original dataset $ \calD $.
	Furthermore, we let $ \Xi_K $ denote the random variable whose law generates the subsets $ \calS $ of candidate splitting variables at all internal nodes in $T_K^{\textnormal{\scalebox{0.8}{CART}}}$, conditional on the subsampled training data $ \calD' $. 
	Notice that the pair of random variables $(\calI',\Xi_K)$ has the same law as $\Theta$.

	In order to prove \prettyref{thm:oracle_forest}, it is necessary to first establish an empirical risk bound akin to \prettyref{thm:training}. 
	Let $ \|\cdot\|_{\calD'}$ and $ \| \cdot \|_{\bt} $ denote the analogues of the empirical norms in \prettyref{eq:emp_norm} and \prettyref{eq:emp_normt}, respectively, but based on the subsampled data $ \calD' $.
	Following the notation in the proof of \prettyref{thm:training}, we let $\calE_K $ denote the excess empirical risk $ \|  y  - \hat g(T_K^{\textnormal{\scalebox{0.8}{CART}}})\|_{\calD'}^2 - \|  y-g\|^2_{\calD'}$.
	Similarly, for a terminal node $ \bt $ of $ T_K^{\textnormal{\scalebox{0.8}{CART}}}$ and additive function $ g(\cdot) \in \calG^1$, we define $ \calE_{K}(\bt) $ to be the within-node excess empirical risk $ \| y-\overline y_{\bt}\|^2_{\bt} - \|  y-g\|^2_{\bt} $, but now based on the subsample $ \calD' $.

	Because of the additional randomness injected into the trees, we proceed by bounding the empirical risk averaged with respect to $ \Xi_K $; that is, we aim to bound $ \mathbb{E}_{\Xi_K}(\|  y  - \hat g(T_K^{\textnormal{\scalebox{0.8}{CART}}})\|_{\calD'}^2) $. Note that due to independence of the subsets $ \calS $ across nodes, any terminal node $\bt$ of $ T_{K-1}^{\textnormal{\scalebox{0.8}{CART}}}$ is conditionally independent of $ \Xi_K $ given $ \Xi_{K-1} $. Thus, we study the $ \Xi_K $ randomness of the tree at depth $ K $ conditional on the randomness in the previous $K-1$ levels of the tree. 
	As with the analysis for \prettyref{thm:training}, we begin with the identity
	\begin{equation}
	\begin{aligned} \label{eq:training_recursion_forest}
& \mathbb{E}_{\Xi_K \mid \Xi_{K-1}}( \|y-\hat g(T_K^{\textnormal{\scalebox{0.8}{CART}}} ) \|^2_{\calD'}) \\ & \qquad = \|y-\hat g(T_{K-1}^{\textnormal{\scalebox{0.8}{CART}}}) \|^2_{\calD'}- \sum_{\bt\in T_{K-1}^{\textnormal{\scalebox{0.8}{CART}}} } \frac{N_{\bt}}{a_N}\mathbb{E}_{\Xi_K \mid \Xi_{K-1}}\big(\max_{(j\in\calS, \; s \in \mathbb{R})}\mathcal{IG}( j, s,\bt)\big),
\end{aligned}
\end{equation}
	which results from taking the expected value of \prettyref{eq:training_recursion} with respect to the conditional distribution of $ \Xi_K $ given $ \Xi_{K-1} $ (again, recognizing that the terminal nodes of $ T_{K-1}^{\textnormal{\scalebox{0.8}{CART}}} $ are conditionally independent of $ \Xi_K $ given $ \Xi_{K-1} $).

	We follow a similar argument as the proof of \prettyref{thm:training},
	 except now in lieu of \prettyref{lmm:gain}, one shows that, for each terminal node $ \bt $ of $ T_{K-1}^{\textnormal{\scalebox{0.8}{CART}}}$,
	\begin{equation} \label{eq:forest_lower_main}
	\mathbb{E}_{\Xi_K|\Xi_{K-1}}\big(\max_{(j\in\calS, \; s \in \mathbb{R})}\mathcal{IG}(j,s, \bt)\big) \geq \frac{m}{p}\frac{ \calE_{K-1}^2(\bt)}{\|g\|^2_{\text{TV}}},
	\end{equation}
	provided $\calE_{K-1}(\bt) > 0 $. 
	To see this, notice that we have the lower bound
	\begin{equation} \label{eq:maximum_lower}
	\begin{aligned}
	\mathbb{E}_{\Xi_K \mid \Xi_{K-1}}\big(\max_{(j\in\calS, \; s \in \mathbb{R})} \mathcal{IG}(j, s, \bt)\big) \geq \mathbb{E}_{\Xi_K \mid \Xi_{K-1}}\big(\mathbf{1}(j_{\bt} \in \calS) \mathcal{IG}(j_{\bt}, s_\bt, \bt)\big) 
	\geq \frac{m}{p}\mathcal{IG}(\bt),
	\end{aligned}
	\end{equation}
	where the last inequality follows from the fact that the probability that a specific coordinate index belongs to $ \calS $ is $ \binom{p-1}{m-1}/\binom{p}{m} =  m/p $. The conclusion then follows directly from \prettyref{lmm:gain} since $ \mathcal{IG}(\bt) \geq \calE_{K-1}^2(\bt)/\|g\|^2_{\text{TV}} $ whenever $ \calE_{K-1}(\bt) > 0 $.

	Proceeding in the same way as before with the proof of \prettyref{lmm:gaind} that produced \prettyref{eq:recursion_plus}, and applying \prettyref{eq:forest_lower_main} to \prettyref{eq:training_recursion_forest}, one can easily establish the inequality
	\begin{equation}
	\label{eq:main_plus}
	\mathbb{E}_{\Xi_K|\Xi_{K-1}}( \calE_K) \leq \calE_{K-1} - \frac{m/p}{\|g\|^2_{\text{TV}}}(\calE^+_{K-1})^2,
	\end{equation}
	where $ \calE_{K-1}^{+} = \sum_{\bt\in T_{K-1}^{\textnormal{\scalebox{0.8}{CART}}}: \calE_{K-1}(\bt) > 0}\frac{N_{\bt}}{a_N} \calE_{K-1}(\bt) \geq \calE_{K-1} $.
	Taking expected values of \prettyref{eq:main_plus} with respect to $ \Xi_{K-1} $ and using, in turn, the law of iterated expectations, i.e., $ \mathbb{E}_{\Xi_{K-1}}(\mathbb{E}_{\Xi_K|\Xi_{K-1}}( \calE_K )) =  \mathbb{E}_{\Xi_K}(\calE_K) $, and Jensen's inequality for the square function, i.e., $ \mathbb{E}_{\Xi_{K-1}}((\calE^+_{K-1})^2) \geq (\mathbb{E}_{\Xi_{K-1}}(\calE^+_{K-1}))^2 $, we have
	\begin{equation}
	\begin{aligned} \label{eq:it_partial}
	\mathbb{E}_{\Xi_K}( \calE_K) & \leq  \mathbb{E}_{\Xi_{K-1}}( \calE_{K-1}) - \frac{m/p}{\|g\|^2_{\text{TV}}} \mathbb{E}_{\Xi_{K-1}}((\calE^+_{K-1})^2) \\ & \leq  \mathbb{E}_{\Xi_{K-1}}( \calE_{K-1}) - \frac{m/p}{\|g\|^2_{\text{TV}}}(\mathbb{E}_{\Xi_{K-1}}(\calE^+_{K-1}))^2.
	\end{aligned}
	\end{equation}

Let us first assume that $\mathbb{E}_{\Xi_{K-1}}(\calE_{K-1})\ge 0$.	
	Since $ \mathbb{E}_{\Xi_{K-1}}(\calE^+_{K-1}) \geq \mathbb{E}_{\Xi_{K-1}}(\calE_{K-1}) \ge 0$ (which follows from $ \calE_{K-1}^{+} \geq \calE_{K-1} $), from \prettyref{eq:it_partial} we finally obtain
	$$
	\mathbb{E}_{\Xi_K}( \calE_K) \leq  \mathbb{E}_{\Xi_{K-1}}(\calE_{K-1})\Big(1-\frac{m}{p}\frac{\mathbb{E}_{\Xi_{K-1}}( \calE_{K-1})}{\|g\|^2_{\text{TV}}}\Big),
	$$
	provided $ \mathbb{E}_{\Xi_{K-1}}( \calE_{K-1}) \geq 0 $.
	Iterating this recursion as with \prettyref{eq:main_recursion} in the proof of \prettyref{thm:training} yields $  \mathbb{E}_{\Xi_K}( \calE_K) \leq  (p/m)\|g\|^2_{\text{TV}}/(K+3) $, or equivalently,
	\begin{equation} \label{eq:training_rf}
	\mathbb{E}_{\Xi_K}(\|  y - \hat g(T_K^{\textnormal{\scalebox{0.8}{CART}}} )\|^2_{\calD'}) \leq \|  y - g \|^2_{\calD'}+\frac{p}{m} \frac{\|g\|^2_{\text{TV}}}{K+3},
	\end{equation}
	for $ K \geq 1 $.
	Now suppose that $\mathbb{E}_{\Xi_{K-1}}(\calE_{K-1})< 0$.	Taking the expectation of \prettyref{eq:training_recursion_forest} with respect to $\Xi_{K-1}$ we see that $  \mathbb{E}_{\Xi_{K}}(\calE_{K})\leq \mathbb{E}_{\Xi_{K-1}}(\calE_{K-1})< 0 $ and hence \prettyref{eq:training_rf} still holds true.
	
	We next turn our attention to modifying the proof of \prettyref{thm:oracle} to accommodate the current setting. The most notable difference is in bounding the probability that $ E_1 \geq 0 $, where $ E_1 $ is defined in \prettyref{eq:E1}. However, this can easily be done in the same manner as before by noting that
	\begin{equation*}
	\begin{aligned}
	& \mathbb{P}_{\calD' \mid \calI'}\big(\mathbb{E}_{\Xi_K}(\| \hat g(T_K^{\textnormal{\scalebox{0.8}{CART}}} ) - g^*\|^2) \geq 2(\mathbb{E}_{\Xi_K}(\|  y -\hat g(T_K^{\textnormal{\scalebox{0.8}{CART}}} )\|^2_{\calD'}) - \|  y - g^*\|_{\calD'}^2) + \alpha + \beta \big) \leq \\ & \qquad\qquad
	\mathbb{P}_{\calD' \mid \calI'}\big(\exists \; g(\cdot) \in \calG_{N}: \|g - g^* \|^2 \geq 2(\|  y - g\|^2_{\calD'}- \|  y - g^*\|_{\calD'}^2) + \alpha + \beta \big),
	\end{aligned}
	\end{equation*}
	since if $ \mathbb{E}_{\Xi_K}(\| \hat g(T_K^{\textnormal{\scalebox{0.8}{CART}}}) - g^*\|^2 - 2\|  y - \hat g(T_K^{\textnormal{\scalebox{0.8}{CART}}})\|^2_{\calD'}+ 2\|  y -g^*\|_{\calD'}^2 - \alpha - \beta) \geq 0 $,
	 then there exists a realization of $ \Xi_K $, yielding a piece-wise constant function $  g(\cdot) $ in $ \calG_N $,
	 for which the inequality $ \| g -g^*\|^2- 2\|  y -  g\|^2_{\calD'}+ 2\|  y -g^*\|_{\calD'}^2 - \alpha - \beta \geq 0 $ also holds.
	Following the same lines as the rest of the proof of \prettyref{thm:oracle}, we have that for all $ K \geq 1 $ and all $ g(\cdot) \in \calG^1$, 
	\begin{equation} \label{eq:forest_upper_main}
	\mathbb{E}_{\Xi_K, \calD' | \calI'}(\|\hat g(T_K^{\textnormal{\scalebox{0.8}{CART}}}) - g^*\|^2) \leq 2\|g- g^*\|^2 +\frac{p}{m} \frac{2\|g\|^2_{\text{TV}}}{K+3} + C_1\frac{2^{K}\log^2(a_N)\log(a_Np)}{a_N},
	\end{equation}
	where $ C_1 $ is the same constant in \prettyref{thm:oracle}. The output of the random forest \prettyref{eq:random_forest} is simply an equally weighted convex combination of the individual tree outputs, each of which is generated according to the law of $ \Theta$. 
	Taking expectations of \prettyref{eq:forest_upper_main} with respect to $ \calI' $ and using Jensen's inequality on the (convex) squared error loss, it follows that
	$$
	\mathbb{E}_{\bTheta, \calD}(\|B^{-1}\textstyle\sum_{b=1}^B\hat g(T_K^{\textnormal{\scalebox{0.8}{CART}}}(\Theta_b))  - g^*\|^2) \leq B^{-1}\textstyle\sum_{b=1}^B\mathbb{E}_{\Theta_b, 
	\calD}(\|\hat g(T_K^{\textnormal{\scalebox{0.8}{CART}}}(\Theta_b)) - g^*\|^2),
	$$
	has the same bound as \prettyref{eq:forest_upper_main},
     which proves \prettyref{thm:oracle_forest}.
\end{proof}

\bibliography{references}
\bibliographystyle{plainnat}

\end{document}